\documentclass[conference]{IEEEtran}
\IEEEoverridecommandlockouts
\usepackage{cite}
\usepackage{amsmath,amssymb,amsfonts}
\usepackage{amsthm}
\DeclareMathOperator*{\argmax}{arg\,max}

\newtheorem{myDef}{Definition}

\usepackage{graphicx}
\usepackage{hyperref}
\usepackage{algorithmicx}
\usepackage{textcomp}
\usepackage[pdftex,dvipsnames]{xcolor}
\usepackage[colorinlistoftodos,prependcaption,textsize=small]{todonotes}
\usepackage{multicol, multirow}
\usepackage{subfigure}
\usepackage{threeparttable}
\usepackage{booktabs}
\usepackage{algorithm}
\usepackage{algpseudocode}
\usepackage{authblk}

\algrenewcommand\algorithmicrequire{\textbf{Input:}}
\algrenewcommand\algorithmicensure{\textbf{Output:}}
\def\BibTeX{{\rm B\kern-.05em{\sc i\kern-.025em b}\kern-.08em
    T\kern-.1667em\lower.7ex\hbox{E}\kern-.125emX}}
\begin{document}

\title{Memorize, Factorize, or be Naïve: Learning Optimal Feature Interaction Methods for CTR Prediction}

\author[1,2]{Fuyuan Lyu* \dag \thanks{\dag This work is done when Fuyuan Lyu worked as an intern at Huawei Noah’s Ark Lab.}}
\author[2]{Xing Tang* \thanks{*Co-first author with equal contribution.}}
\author[2]{Huifeng Guo}
\author[2]{Ruiming Tang\ddag \thanks{\ddag  Corresponding authors.}}
\author[2]{\\Xiuqiang He} 
\author[3]{Rui Zhang\ddag}
\author[1]{Xue Liu}

\affil[1]{School of Computer Science, McGill University}
\affil[2]{Huawei Noah’s Ark Lab} 
\affil[3]{www.ruizhang.info}
\affil[ ]{\text{fuyuan.lyu@mail.mcgill.ca}, \text{xueliu@cs.mcgill.ca}, \text{rayteam@yeah.net}}
\affil[ ]{\text{\{xing.tang, huifeng.guo, tangruiming, hexiuqiang1\}@huawei.com}}

\maketitle

\newcommand{\ruiming}[1]{{\bf \color{red} [[Ruiming says ``#1'']]}}
\newcommand{\change}[1]{#1}
\newcommand{\info}[1]{}

\begin{abstract}
Click-through rate prediction is one of the core tasks in commercial recommender systems. It aims to predict the probability of a user clicking a particular item given user and item features. As feature interactions bring in non-linearity, they are widely adopted to improve the performance of CTR prediction models. Therefore, effectively modelling feature interactions has attracted much attention in both the research and industry field. The current approaches can generally be categorized into three classes: (\romannumeral1) \emph{naïve} methods, which do not model feature interactions and only use original features; (\romannumeral2) \emph{memorized} methods, which memorize feature interactions by explicitly viewing them as new features and assigning trainable embeddings; (\romannumeral3) \emph{factorized} methods, which learn latent vectors for original features and implicitly model feature interactions through factorization functions. 
Studies have shown that modelling feature interactions by one of these methods alone are suboptimal due to the unique characteristics of different feature interactions. To address this issue, we first propose a general framework called \textit{OptInter} which finds the most suitable modelling method for each feature interaction. Different state-of-the-art deep CTR models can be viewed as instances of \textit{OptInter}. To realize the functionality of \textit{OptInter}, we also introduce a learning algorithm that automatically searches for the optimal modelling method. We conduct extensive experiments on four large datasets, including three public and one private. Experimental results demonstrate the effectiveness of \textit{OptInter}. Because our \textit{OptInter} finds the optimal modelling method for each feature interaction, our experiments show that \textit{OptInter} improves the best performed state-of-the-art baseline deep CTR models by up to 2.21\%. Compared to the \textit{memorized} method, which also outperforms baselines, we reduce up to 91\% parameters. In addition, we conduct several ablation studies to investigate the influence of different components of \textit{OptInter}. Finally, we provide interpretable discussions on the results of \textit{OptInter}.

\end{abstract}
\begin{IEEEkeywords}
Click-through Rate Prediction, Feature Interaction, Recommendation, Neural Architecture Search
\end{IEEEkeywords}

\section{Introduction}
\label{sec:intro}





The Click-through rate (CTR) prediction task is crucial for recommender systems, which aims to predict the probability of a certain user clicking on a recommended item (e.g. movie, advertisement)\cite{Wide_Deep,DeepFM,PNN16,PNN19}. Many recommendations can therefore be performed based on the result of CTR prediction. For instance, to maximize the number of clicks, the items returned to a user can be ranked by predicted CTR (pCTR). 

Due to the powerful featur  e representation learning ability, the mainstream of CTR prediction research is dominated by deep learning models. 
As an important research direction to improve deep CTR models, many methods of modelling effective feature interactions are proposed, such as~\cite{FNN,DeepFM,Wide_Deep,PNN16,PNN19}. 

\begin{figure*}[!htbp]
\centering
\subfigure[FNN]{
\begin{minipage}[t]{0.27\textwidth}
\centering
\includegraphics[width=\textwidth]{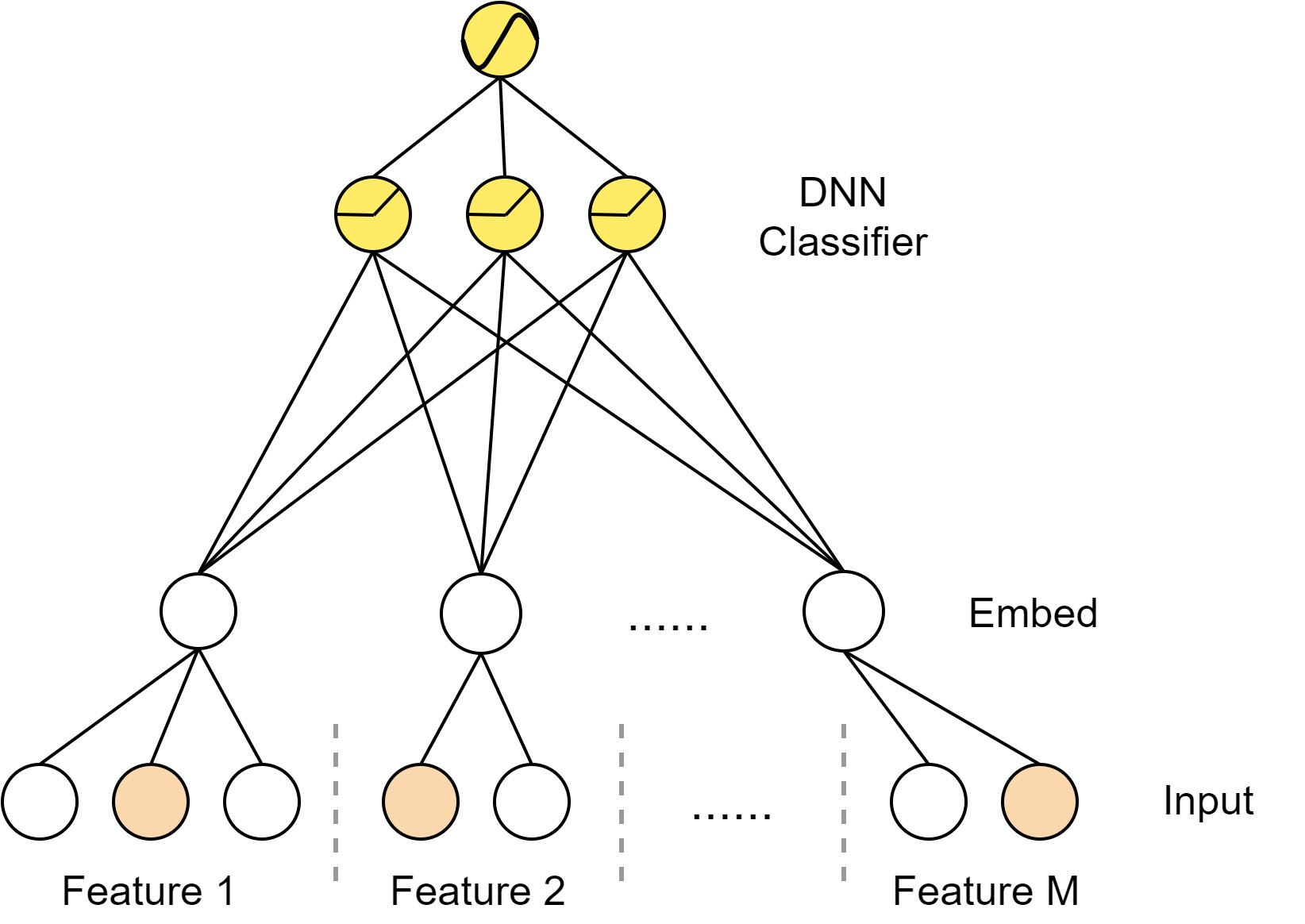}
\label{fig:FNN}
\end{minipage}
}
\subfigure[Wide\&Deep]{
\begin{minipage}[t]{0.4\textwidth}
\centering
\includegraphics[width=\textwidth]{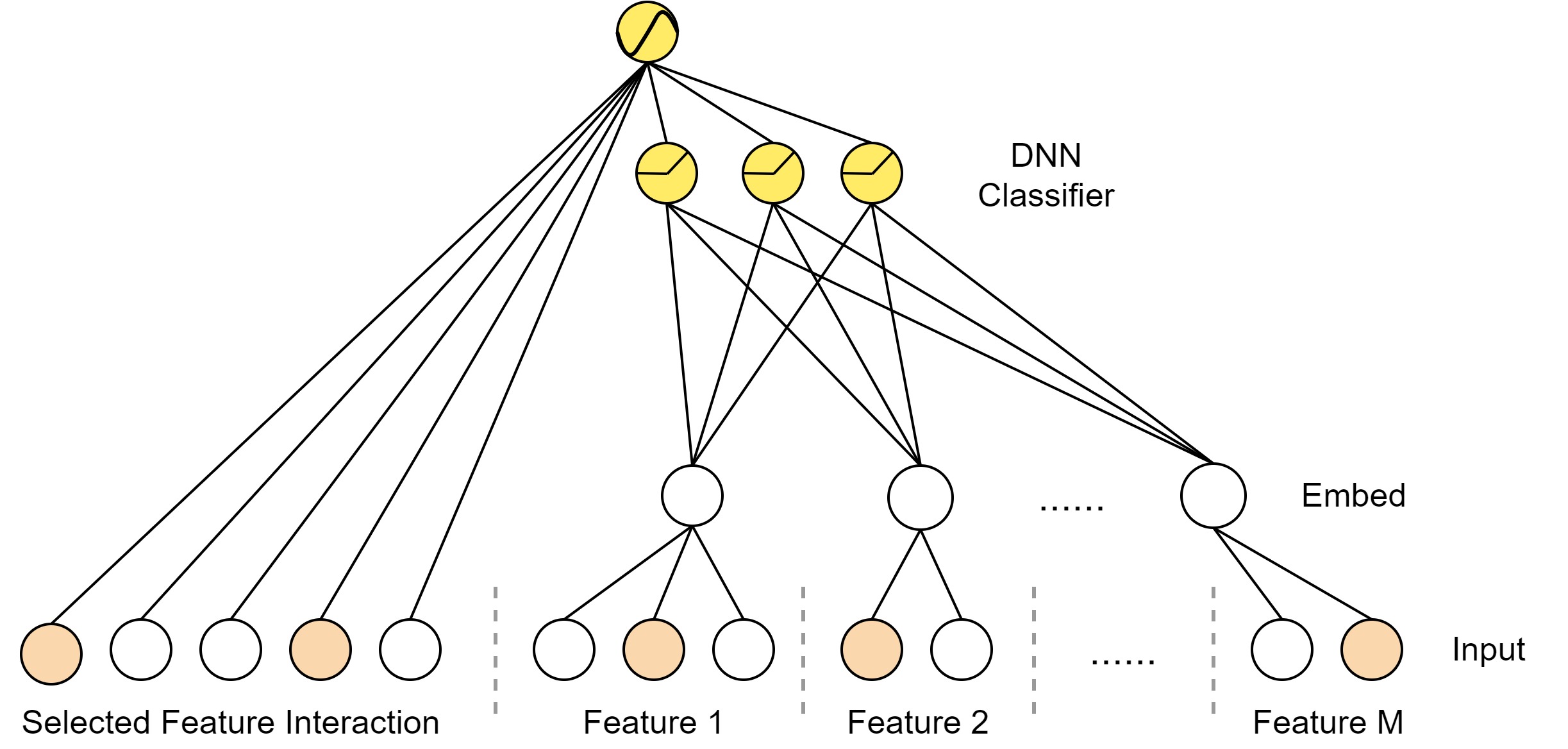}
\label{fig:Wide&Deep}
\end{minipage}
}
\subfigure[IPNN]{
\begin{minipage}[t]{0.27\textwidth}
\centering
\includegraphics[width=\textwidth]{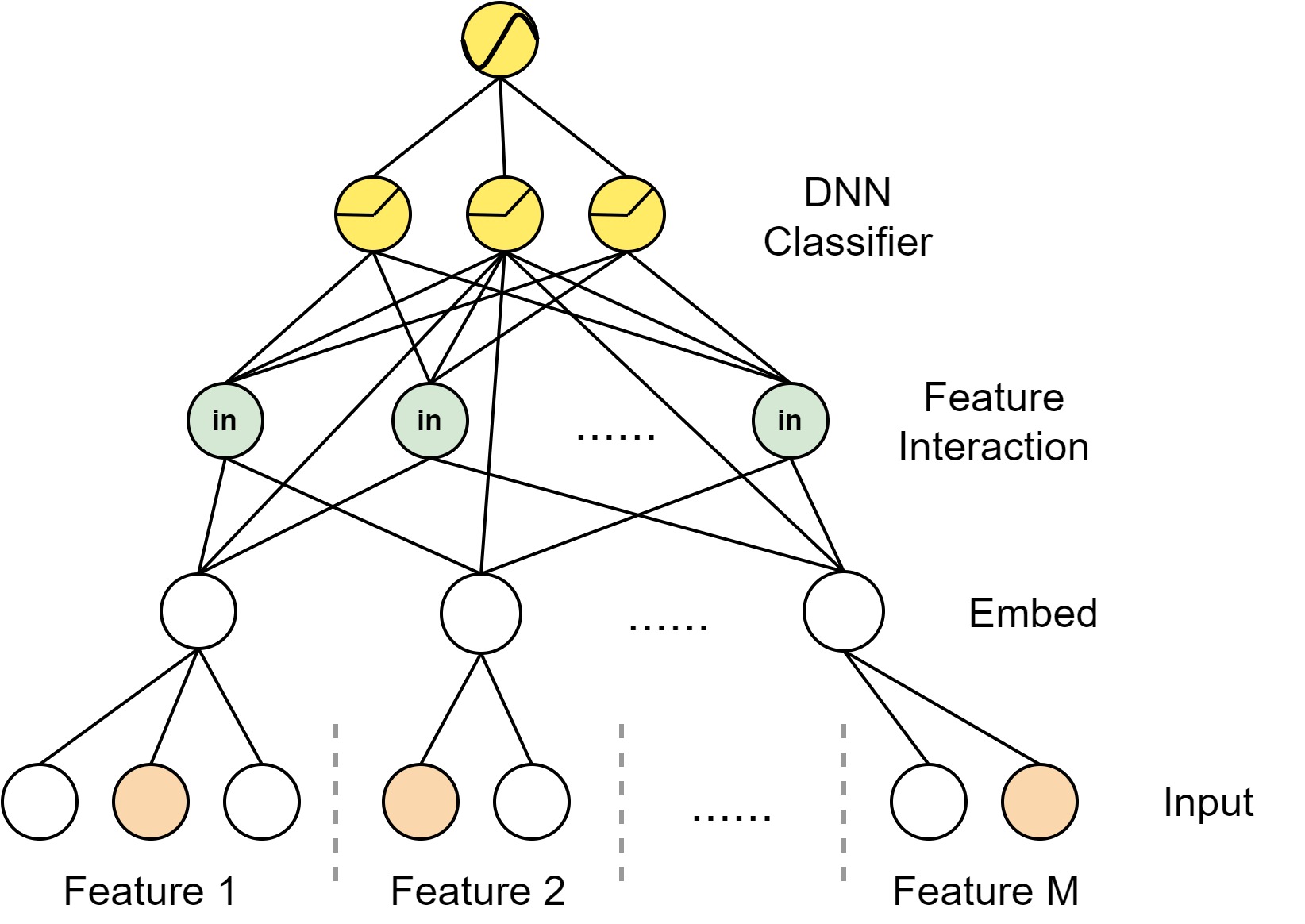}
\label{fig:IPNN}
\end{minipage}
}
\caption{Examples of modeling feature interactions: Naïve, Memorized and Factorized}
\label{fig:example}
\end{figure*}

The simplest way of modelling feature interactions is feeding original features in Multi-layer Perceptron (MLP). Shown as an example in Figure ~\ref{fig:FNN}, FNN~\cite{FNN} directly feeds original features into MLP and relies on the capability of MLP to model feature interactions. The universal approximation rule has proved that MLP can mimic arbitrary functions given enough data and computation power~\cite{Universal}. However, it is challenging for MLP to model low-rank feature interactions solely based on original features~\cite{Latent-Cross}. Such a way of modelling feature interactions by MLP directly is referred to as \emph{naïve} methods in our paper.

Another alternative to model feature interactions is memorizing them explicitly as new features, named as \emph{memorized} methods. These methods~\cite{Poly-2,Wide_Deep} which memorize all second-order feature interactions as new features and feed them in shallow models (as shown in Figure~\ref{fig:Wide&Deep}, feature interactions are treated as individual features and fed into the wide component), achieve superior performance than \emph{naïve} methods. The reason is that some feature interactions are served as strong signals such that memorizing them as new features makes correlated patterns much easier to capture. \change{However, \emph{memorized} methods are prone to overfitting as the new features (generated by feature interactions) are more sparse and have lower frequency than original features in the training set.}

\info{R1O1: We move certain discussion from the introduction section into the related work section.}

\change{The final method is to model feature interactions via a factorization function, named as \emph{factorized} method. It is originated from Factorization Machine (FM)~\cite{FM} and its variants~\cite{FFM,FwFM,FM2}. \emph{Factorized} methods implicitly model all second-order feature interactions by learning latent vectors of original features and aggregating them using a specific function (e.g., inner-product)~\cite{FM,FFM} or learnable factorization function~\cite{FwFM,FM2}, as shown in Figure~\ref{fig:IPNN}. \emph{Factorized} methods alleviate the feature sparsity issue and are widely adopted by the mainstream deep CTR models~\cite{DeepFM,PNN16,PNN19}. However, the latent vectors are used in both representation learning and feature interaction modelling, which tends to conflict with each other.}

The above three methods (namely, \emph{naïve, memorized, factorized} methods) model all possible feature interactions in an identical way. 
However, as stated in~\cite{AutoFis}, modelling feature interactions in the same way may lead to a suboptimal solution because the characteristics (e.g., complexity) of each feature interaction may not be identical.
Hence, AutoML methods~\cite{DARTS,Large-Scale-Evolution} are introduced to find an appropriate modelling method for each feature interaction~\cite{AutoFis,AutoFeature}. For instance, AutoFIS~\cite{AutoFis} aims to select which feature interactions should be factorized and which ones should be ignored. In other words, AutoFIS makes a choice between \emph{factorized} and \emph{naïve} adaptively for each individual feature interaction, while the option of \emph{memorized} is neglected.

With the limitations of prior research observed, a data-driven strategy to automatically find an optimal method from \emph{naïve, memorized, factorized} methods for each feature interaction is required. This motivates us to propose a general framework called \textit{OptInter}. For each feature interaction, \textit{OptInter} selects the optimal modelling method from \emph{naïve, memorized, factorized} methods adaptively and automatically. 
Inspired by DARTS~\cite{DARTS} and previous works~\cite{AutoFis,AutoFeature},
to efficiently search the optimal method for each feature interaction, we devise a two-stage learning algorithm. 
In the \textit{search} stage, we aim to select the optimal method for each feature interaction from the search space \emph{naïve, memorized, factorized}, where the selection is treated as a neural architecture search problem. 
However, such a selection is discrete and makes the overall framework not differentiable. Therefore, instead of searching over three candidate modelling methods, we relax the choices to be continuous by approximating with Gumbel-softmax tricks\cite{Gumbel-Softmax} via a set of architecture parameters (one for each modelling method with respect to a feature interaction). Then, the architecture parameters can be learned by gradient descent, which is jointly optimized with neural network weights. In the \textit{re-train} stage, we select the modelling method with the largest probability for each feature interaction and re-train the model from scratch. The neural network weights obtained from the \textit{search} stage are discarded to avoid the influence of suboptimal modelling methods. 

Extensive experiments are conducted on four large-scale datasets, including three public datasets and one private dataset. Experimental results demonstrate \textit{OptInter} consistently performs well on all datasets. Specifically, \emph{memorized} method improves the best performed deep CTR model by 0.1\%-2.2\% in terms of the AUC score with the help of 18 times more parameters. Moreover, compared to the \emph{memorized} method, \textit{OptInter} achieves further improvement of AUC by 0.01\%-0.25\%  while reduces about 18\%-91\% parameters. The results demonstrate the effectiveness of introducing \emph{memorized} method in our search space and the efficiency of \textit{OptInter} by selecting suitable modelling methods for individual feature interactions. Our ablation studies also show that our proposed search algorithm yields a more effective architecture than other search algorithms. Last but not least, we analyze the search architecture of \textit{OptInter} from the perspective of information theory and provide interpretability. To sum up, the main contributions of this paper are:
\begin{enumerate}
    \item We propose a novel deep CTR prediction framework \textit{OptInter} including \emph{naïve, memorized, factorized} feature interaction methods. To our best knowledge, \textit{OptInter} is the first work to introduce \emph{memorized} method in deep CTR models. Moreover, some mainstream deep CTR methods can be viewed as instances of \textit{OptInter}.
    \item As a part of \textit{OptInter}, we propose a two-stage learning algorithm to select the optimal method for each feature interaction automatically. In the search stage, \textit{OptInter} can learn the relative importance of each feature interaction method via architecture parameters. In the re-train stage, with the resulting optimal methods, we re-train the model from scratch to guarantee the neural network is not influenced by suboptimal methods.
    \item Comprehensive experiments are conducted on three public datasets and one private dataset and show that \textit{OptInter} outperforms the state-of-the-art deep CTR prediction models. The results demonstrate that \textit{OptInter} is both effective and efficient. 
\end{enumerate}

\section{Methodology}

In this section, we first formulate the problem of CTR prediction task and feature interaction modelling methods in Section \ref{sec:problem}. Then we describe the proposed framework \textit{OptInter} in Section \ref{sec:model}. Finally, we elaborate details of the learning algorithm for \textit{OptInter} in Section \ref{sec:learning}. To make our framework easier to understand, we list all the notations used in Table \ref{Table:notation}.

\subsection{Problem Formulation}
\label{sec:problem}

\subsubsection{CTR Prediction}
A dataset for training CTR models consists of instances $(\mathbf{X}^{o},y)$, where $y \in \{0,1\}$ is the ground truth label and $\mathbf{X}^{o}$ is a multi-field data instance including $M$ original features
\begin{equation}
    \mathbf{X}^{o} = [\mathbf{x}^{o}_{1}, \mathbf{x}^{o}_{2}, ..., \mathbf{x}^{o}_{M}],
\end{equation}
where $\mathbf{x}^{o}_i$ is the one-hot encoded vector for the feature value in the $i$-th original feature. The problem of CTR prediction is to predict the probability of a user clicking a certain item according to original features $\mathbf{X}^{o}$. Formally, a machine learning model to estimate the probability is defined as follows,

\begin{equation}
    \mathbf{P}(y=1 | \mathbf{X}^o) = f(\mathbf{X}^o;\Theta),
    \label{eq:form}
\end{equation}
where $f$ is the model, ${\Theta}$ indicates all the model parameters, and $\mathbf{P}$ is the conditional probability.

\begin{table}[htbp]   
\renewcommand\arraystretch{1.15}
\centering
\caption{Selected notations in this paper}  
\begin{tabular}{l|l}    
\hline
\textbf{Notations} & \textbf{Descriptions} \\    
\hline
$y$ & ground truth label\\
$\hat{y}$ & predicted result \\
$\mathbf{X}^{o}$ & original features\\
$\mathbf{X}^{m}$ &  cross-product transformed features \\
$\mathbf{x}^{o}_i$ & original feature $i$ \\
$\mathbf{e}^{o}_i$ & embedding for original feature $i$ \\
$\mathbf{e}^{f}_{(i,j)}$ & factorized embedding for $(\mathbf{x}^o_i,\mathbf{x}^o_j)$ \\
$\mathbf{x}^{m}_{(i,j)}$ & cross-product transformed features for $(\mathbf{x}^o_i,\mathbf{x}^o_j)$ \\
$\mathbf{e}^{m}_{(i,j)}$ & memorized embedding for $(\mathbf{x}^o_i,\mathbf{x}^o_j)$ \\
$\mathbf{e}^{n}$ & null embedding \\
$\mathbf{e}^{b}_{(i,j)}$ & optimal embedding for $(\mathbf{x}^o_i,\mathbf{x}^o_j)$ \\
$M$ & number of original feature \\
$\mathcal{K}$ & searchable modelling space \\
$\mathcal{D}$ & dataset \\
$\times$ & cross product \\
$\Theta$ & model parameter \\
$\alpha$ & architecture parameter \\
\hline
\end{tabular}
\label{Table:notation}
\end{table}

\subsubsection{Feature Interaction}
\label{sec:feature_interaction}
Feature interactions capture the correlation between different features and induce non-linearity to the model. As shown in existing works, it is crucial to model feature interactions to boost the model performance~\cite{PNN16,PNN19,DeepFM,Wide_Deep,FNN}. 

In this section, we formally define the feature interaction and introduce three modelling methods.

\begin{myDef}
\label{featuredef}

A $h$-th order ($1 \le h \le M$) feature interaction $\mathcal{H}$ is defined as a multivariate group $(\mathbf{x}^{o}_{c_1}, \mathbf{x}^{o}_{c_2}, ..., \mathbf{x}^{o}_{c_h})$, where each feature $\mathbf{x}^{o}_{c_i}$ is selected from original feature $\mathbf{X}^{o}$. 

Generally, there are three methods to define feature interaction: 

(\romannumeral1) \emph{Factorized} method: Given the latent vectors for original features $\mathbf{E}^{o}$, $\mathcal{H}$ can be modelled as\cite{AutoPI}
\begin{equation}
\begin{split}
    \mathbf{e}^{f}_{\mathcal{H}} &= o^{(h-1)}(... ( o^{(1)}(\mathbf{e}^{o}_{c_1}, \mathbf{e}^{o}_{c_2})), ... , \mathbf{e}^{o}_{c_h}), \\
    \mathbf{e}^{o}_{c_i} &= \mathbf{E}^{o} \mathbf{x}^{o}_{c_i}.
\end{split}
\end{equation}
Namely, the factorized embedding $\mathbf{e}^{f}_{\mathcal{H}}$ is generated by the utilizing $h-1$ operators $o^{(1)}(\cdot)$, $o^{(2)}(\cdot)$, .., $o^{(h-1)}(\cdot)$ in order to aggregate $h$ latent vectors $\mathbf{e}^{o}_{c_1}$, $\mathbf{e}^{o}_{c_2}$, ..., $\mathbf{e}^{o}_{c_h}$. Here the operators could be FM, inner product or etc.

(\romannumeral2) \emph{Memorized} method: Memorized embedding $\mathbf{e}^{m}_{\mathcal{H}}$ can be defined as\cite{FIVES} 
\begin{equation}
\begin{split}
\label{memorized}
    \mathbf{e}^{m}_{\mathcal{H}} &= \mathbf{E}^{m} \mathbf{x}^{m}_{\mathcal{H}}, \\
    \mathbf{x}^{m}_{\mathcal{H}} &= \text{onehot}(\mathbf{\bar{x}}^{o}_{c_1} \times \mathbf{\bar{x}}^{o}_{c_2}\times ... \times \mathbf{\bar{x}}^{o}_{c_h}),
\end{split}
\end{equation}
where $\mathbf{\bar{x}}^{o}_{c_i}$ is zero-padded original feature $\mathbf{x}^{o}_{c_i}$ (zero-padding is introduced to align the dimension), $\times$ is the cross-product operation, $\mathbf{x}^{m}_{\mathcal{H}}$ is the cross-product transformed feature for $\mathcal{H}$, $\mathbf{E}^{m}$ is the embedding table for cross-product transformed features.

(\romannumeral3) \emph{Naïve} method: the naïve embedding $\mathbf{e}^n_{\mathcal{H}} = \mathbf{e}^n$ can be defined as a zero vector with arbitrary length. Note that for different $\mathcal{H}$, the naïve embedding $\mathbf{e}^n$ remains the same.

\end{myDef}

\subsection{OptInter Framework}
\label{sec:model}

\begin{figure}[!htbp]
\centering
\includegraphics[width=0.45\textwidth]{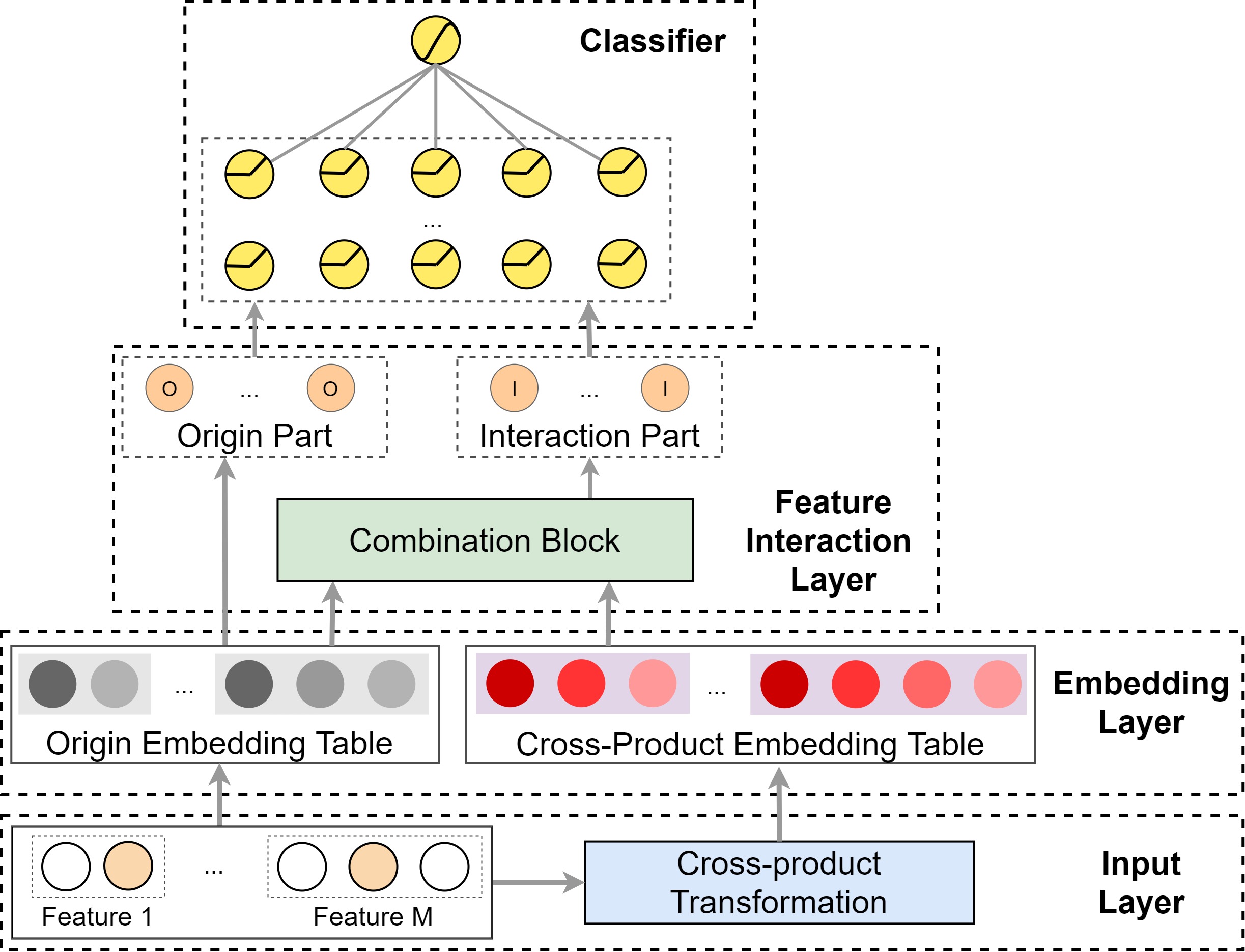}
\caption{The OptInter CTR Framework}
\label{fig:illustration}
\end{figure}

In this section, we elaborate the \textit{OptInter} framework in detail, which is presented in Figure \ref{fig:illustration}. The \textit{OptInter} framework consists of four components: (\romannumeral1) the input layer (Section~\ref{input}), (\romannumeral2) the embedding layer (Section~\ref{embedding}), (\romannumeral3) the feature interaction layer (Section~\ref{feature}) and (\romannumeral4) the classifier (Section~\ref{classifier}). The feature interaction layer plays a central part in \textit{OptInter} and greatly influences the performance, where the combination block searches the optimal method for each individual feature interaction.

In \textit{OptInter}, Equation \ref{eq:form} is specified to estimate the probability for input $\mathbf{X^o}$ as follows,
\begin{equation}
\begin{split}
    \hat{y} = f(\mathbf{X}^o|\Theta, \alpha),
\end{split}
\end{equation}
where $f(\cdot)$ indicates the \textit{OptInter} framework, $\Theta$ indicates all the neural network parameters and $\alpha$ indicates the architecture parameters. Such architecture parameters in combination block decide which modelling method to choose for each feature interaction.

\subsubsection{Input Layer}
\label{input}

In \textit{OptInter}, the input layer contains a cross-product transformation block, which takes original features $\mathbf{X}^{o}$ as input and generates one-hot encoded representation $\mathbf{X}^{m}$ according to Equation \ref{memorized}. Note that for $M$ original features, there are $C^h_M$ $h$-th order feature interactions. In \textit{OptInter}, we only consider the second-order feature interactions (namely, $h=2$) for two reasons: (\romannumeral1) second-order feature interactions have been demonstrated to be the most important for prediction~\cite{AutoFeature} and (\romannumeral2) modelling higher-order feature interactions exponentially increases the size of the embedding table and makes the current hardware difficult to train such large models. 
Note that although we only consider second-order feature interactions in this paper, our methods could easily be extended to higher-order. The embeddings of all the second-order feature interactions are concatenated to form

\begin{equation}
    \mathbf{X}^{m} = [\mathbf{x}^{m}_{(1,2)}, \mathbf{x}^{m}_{(1,3)}, ..., \mathbf{x}^{m}_{(M-1,M)}].
\end{equation}

The final input includes both original feature $\mathbf{X}^{o}$ and cross-product transformed features $\mathbf{X}^{m}$ for feature interactions.

\subsubsection{Embedding Layer}
\label{embedding}

The embedding layer is to transform the one-hot encoded features into continuous embeddings. After going through input layer, the features contains two parts: (\romannumeral1) original features $\mathbf{X}^{o}$ and (\romannumeral2) cross-product transformed features $\mathbf{X}^{m}$. Both $\mathbf{X}^{o}$ and $\mathbf{X}^{m}$ are one-hot encoded and multi-field. All the features are in categorical form, where features in the numerical form are usually transformed into categorical form by bucketing~\cite{DeepFM,PNN16}.
For univalent features (e.g., ``Gender=Male"), we embed the one-hot encoding of each feature separately to a continuous vector using a linear embedding, i.e., the embedding $\mathbf{e}^{o}_i$ of feature $i$ is $\mathbf{e}^{o}_i = \mathbf{E}^{o} \mathbf{x}^{o}_i$, where $\mathbf{E}^{o}$ is the embedding table for original features. For multivalent features (e.g., ``Interest=Football, Basketball"), we keep the same procedure as \cite{PNN16,PNN19,DeepFM} where all the embeddings of individual feature values are aggregated by mean pooling. The embeddings of the original features are concatenated to form
\begin{equation}
    \mathbf{e}^{o} = [\mathbf{e}^{o}_1, \mathbf{e}^{o}_2, ..., \mathbf{e}^{o}_M].
\end{equation}

The cross-product transformed features $\mathbf{X}^{m}$ for all feature interactions are also embedded to a continuous vector in the same way. For the cross-product transformed feature between original feature $\mathbf{x}^o_i$ and $\mathbf{x}^o_j$, its embedding $\mathbf{e}^{m}_{(i,j)} = \mathbf{E}^{m} \mathbf{x}^{m}_{(i,j)}$, where $\mathbf{E}^{m}$ is the embedding table for cross-product transformed features. All $\mathbf{e}^{m}_{(i,j)}$s are concatenated to form

\begin{equation}
    \mathbf{e}^{m} = [\mathbf{e}^{m}_{(1,2)}, \mathbf{e}^{m}_{(1,3)}, ..., \mathbf{e}^{m}_{(M-1,M)}].
\end{equation}

\subsubsection{Feature Interaction Layer}
\label{feature}
The feature interaction layer is the nucleus of \textit{OptInter}. As observed in Figure \ref{fig:illustration}, we introduce a combination block to search for the optimal modelling method for each feature interaction. The combination block takes both the original feature embedding $\mathbf{e}^{o}$ and the cross-product transformed feature embedding $\mathbf{e}^{m}$ as inputs and generates embeddings $\mathbf{e}^{b}$ for all the feature interactions with the methods searched by \textit{OptInter}. For each feature interaction, we search its optimal modelling method from (\romannumeral1) \emph{memorized} method with the \emph{memorized} cross-product embedding $\mathbf{e}^{m}$; (\romannumeral2) \emph{factorized} method with the original feature embedding $\mathbf{e}^{o}$ and factorization function; and (\romannumeral3) \emph{naïve} method which indicates this feature interaction is useless and does not need to modelling. We will elaborate the learning algorithm used to search the optimal method for each feature interaction in Section \ref{sec:learning}.

\subsubsection{Classifier}
\label{classifier}
In the feature interaction layer, $\mathbf{e}^{o}$ and $\mathbf{e}^{b}$ are concatenated into a single vector $\mathbf{e}=[\mathbf{e}^{o}, \mathbf{e}^{b}]$. Similar to previous works\cite{PNN16,DeepFM,PNN19}, such vector $\mathbf{e}$ is fed into MLP, in which layer normalization and ReLU are applied. One such layer in MLP is defined as
\begin{equation}
\begin{split}
    & \mathbf{a^{(l+1)}} = \text{LN}(\text{relu}(\mathbf{W^{(l)}}\mathbf{a^{(l)}} + \mathbf{b^{(l)}})), \\
    & \mathbf{a^0} = \mathbf{e},
\end{split}
\end{equation}
where $\mathbf{a^{(l)}}$, $\mathbf{W^{(l)}}$ and $\mathbf{b^{(l)}}$ are the input, model weight, and bias of the $l$-th layer. Activation ReLU is defined as
\begin{equation}
    \text{relu}(z)=\max(0,z),
\end{equation}
and layer normalization is defined as \cite{LayerNorm} 
\begin{equation}
    \textbf{LN}(z) = \frac{z - \mathcal{E}[z]}{\sqrt{\text{Var}[x]+\epsilon}} * \gamma + \beta.
\end{equation}

Note that $\text{MLP}(\mathbf{a}^{(0)})=\mathbf{a}^{(h)}$, where $h$ is the depth of MLP. Finally, the output of MLP, $\mathbf{a}^{(h)}$, is fed into a sigmoid unit, in order to re-scale the prediction value to a probability. Formally, the final predicted result is 
\begin{equation}
    \hat{y} = \text{sigmoid}(\mathbf{a}^{(h)}) = \frac{1}{1+e^{-\mathbf{a}^{(h)}}} \in (0,1),
    \label{eq:y_hat}
\end{equation}
which indicates the probability of a specific user clicking on the item. To train our CTR prediction model, we use the cross-entropy loss (i.e., log-loss) function

\begin{equation}
\begin{split}
    \mathcal{L}(\mathcal{D}|\Theta, \alpha) &= -\frac{1}{|\mathcal{D}|} \sum_{(x, y) \in \mathcal{D}} \text{CE}(y, \hat{y}), \\
    \text{CE}(y, \hat{y}) &= y \log(\hat{y}) + (1-y) \log(1-\hat{y}),
\end{split}
\label{eq:loss}
\end{equation}
where $\mathcal{D}$ indicates the training dataset, $\Theta$ includes all the neural network weights $\{\mathbf{W^{(l)}}, \mathbf{b^{(l)}} | 1 \le l \le h \}$ and embedding tables $\{\mathbf{E}^o,\mathbf{E}^m\}$, $\alpha$ is the set of architecture parameters used to search the optimal feature interaction. $\alpha$ will be discussed in Section~\ref{sec:learning}. 

\subsection{Learning Algorithm for OptInter}
\label{sec:learning}

To find the optimal feature interaction method, we need to define a search space and devise an efficient search algorithm. In this section, we first define our search space consisting of \emph{memorized}, \emph{factorized}, and \emph{naïve} methods for each feature interaction. Then we introduce our search algorithm to find the optimal methods efficiently. At last, we present the re-train stage for final training.

\subsubsection{Search Space}

\begin{figure}[!htbp]
\centering
\includegraphics[width=0.45\textwidth]{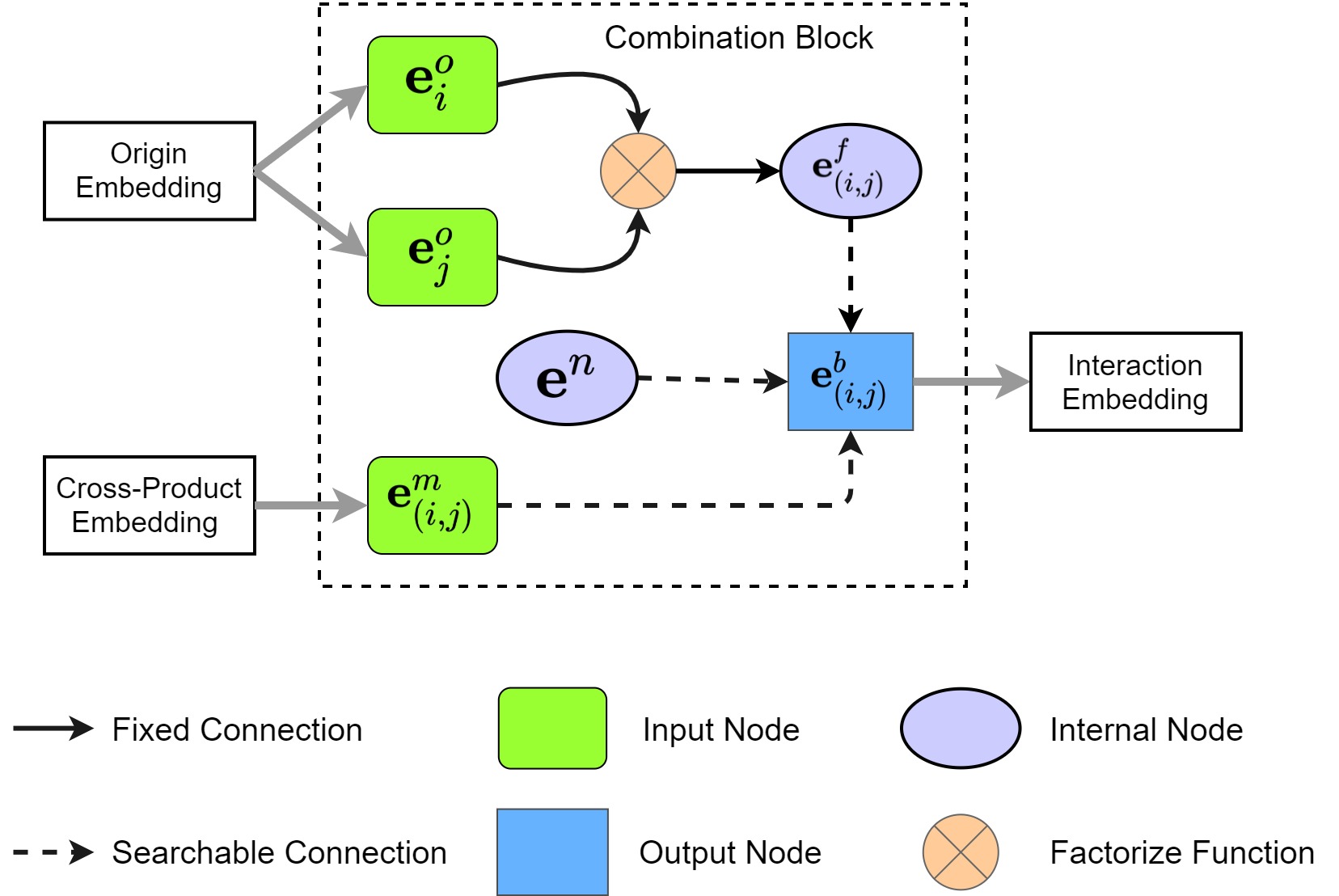}
\caption{Detailed illustration figure about combination block}
\label{fig:gate} 
\end{figure}

As stated in Section~\ref{model}, we categorize feature interaction methods into three classes: (\romannumeral1) \emph{memorized} methods treat each feature interaction as a new feature and explicitly assign trainable weights or embedding. (\romannumeral2) \emph{factorized} methods model a feature interaction via factorization methods on latent vectors of original features. (\romannumeral3) \emph{naïve} methods feed the original features into MLP to model their interactions. Our search space consists of these three methods to model feature interaction. Note that there exist a wide variety of factorization functions, such as \textit{Hadamard Product} $\otimes$, \textit{Pointwise-Addition} $\oplus$ and \textit{Generalized-Product} $\boxtimes$.
In this paper, as we focus on searching the optimal modelling method (namely, \emph{memorized}, \emph{factorized} and \emph{naïve}) for each feature interaction, we take \textit{Hadamard Product} $\otimes$ as the representative for \emph{factorized} methods, instead of considering all possible product operations. Our framework can be extended easily to taking multiple operations into account as \emph{factorized} methods. \textit{Hadamard Product} is element-wise multiplication over each element in the vectors. Formally, the \textit{Hadamard Product} for feature $\mathbf{e}_i$ and $\mathbf{e}_j$ is
\begin{equation}
    \mathbf{e}^{f}_{(i,j)} = \mathbf{e}^{o}_i \otimes \mathbf{e}^{o}_j = [e^1_i \times e^1_j, e^2_i \times e^2_j, ..., e^s_i \times e^s_j],
\end{equation}
where $e^t_i$ is the $t$-th element of the embedding for feature $i$, $s$ is the length of embedding. For each feature interation, we make the choice from 3 options. There are $C_M^2 = M(M-1)/2$ second-order feature interactions in total. Therefore the total search space of \textit{OptInter} is $\mathcal{O}(3^{M(M-1)/2})$, which is an incredibly huge space to search over.

The combination block is visualized in Figure \ref{fig:gate}. The final output $\mathbf{e}^{b}_{(i,j)}$ is chosen from: \emph{memorized} cross-product embedding $\mathbf{e}^{m}_{(i,j)}$, \emph{factorized}  embedding $\mathbf{e}^{f}_{(i,j)}$ or \emph{naïve} embedding $\mathbf{e}^{n}$ (which is actually empty embedding)

\begin{equation}
    \mathbf{e}^{b}_{(i,j)} \in \{\mathbf{e}^{m}_{(i,j)}, \mathbf{e}^{f}_{(i,j)}, \mathbf{e}^{n}\}.
\end{equation}

\subsubsection{Search Algorithm}

\begin{algorithm}
	\caption{The Optimization of Search Stage} 
    \label{alg:search}
	\begin{algorithmic}[1]
		\Require Training dataset $\mathcal{D}$ consists of original features $(x_1, .., x_M)$ and ground-truth labels $y$
        \Ensure the searched optimal architecture parameter $\alpha^{*}$
        \While {not converge}
            \State Sample a mini-batch of training data $\mathcal{D}_{\text{train}}$
            \State Generate the predicted labels $\hat{y}$ via \textit{OptInter} with \Statex \qquad model parameter $\Theta$ and architecture parameter $\alpha$ 
            \Statex \qquad given Equation \ref{eq:y_hat}
            \State Calculate the cross-entropy loss $\mathcal{L}(\mathcal{D}_{\text{train}}|\Theta, \alpha)$ over 
            \Statex \qquad the mini-batch given Equation \ref{eq:loss}
            \State Update model parameter $\Theta$ by descending gradient 
            \Statex \qquad $\bigtriangledown_{\Theta} \mathcal{L}(\mathcal{D}_{\text{train}}|\Theta, \alpha)$
            \State Update architecture parameter $\alpha$ by descending 
            \Statex \qquad gradient $\bigtriangledown_{\alpha} \mathcal{L}(\mathcal{D}_{\text{train}}|\Theta, \alpha)$
        \EndWhile
	\end{algorithmic}
\end{algorithm}

In this section, we propose a search algorithm for exploring the huge search space efficiently. Instead of searching over a discrete (categorical) set of selections on the candidate methods, which will make the whole framework not end-to-end differentiable, we approximate the discrete sampling via introducing the Gumbel-softmax operation\cite{Gumbel-Softmax} in this work. The Gumbel-softmax operation provides a differentiable sampling, which makes the architecture parameters \change{learnable by Adam optimizer}.

To be specific, suppose architecture parameters $\{\alpha_{(i,j)}^k | k \in \mathcal{K} \}$ are the class probability over different feature interaction methods, $\mathcal{K}$ indicates the searchable space over feature interaction methods (namely, \emph{memorized}, \emph{factorized}, \emph{naïve}). Then a discrete selection $z$ can be drawn via the gumbel-softmax trick\cite{Gumbel-Softmax-dist} as

\begin{equation}
\begin{split}
    & z = \text{onehot} \left(\argmax_{k \in \mathcal{K}} [ \log \alpha^k_{(i,j)} + g_{(i,j)} ] \right), \\
    & g_{(i,j)} = -\log(-\log(u_{(i,j)})), \\
    & u_{(i,j)} \sim \text{Uniform}(0,1).
\end{split}
\end{equation}

The \textit{gumbel noise} $g_{(i,j)}$ is i.i.d. sampled, which aims to perturb the log term $\log \alpha_{(i,j)}^k$ and makes the argmax operation \change{equivalent} to drawing a sample by $\{\alpha_{(i,j)}^k | k \in \mathcal{K}\}$ weights. However, the argmax operation makes this trick non-differentiable. To deal with this problem, we replace the argmax operation with the softmax function
\begin{equation}
    p^k_{(i,j)} = \frac{\exp(\frac{\log(\alpha^k_{(i,j)})}{\tau})}{\sum_{k \in \mathcal{K}}\exp(\frac{\log(\alpha^k_{(i,j)})}{\tau})},
\end{equation}
where $\tau$ is the temperature parameter to control the smoothness of the Gumbel-softmax operation. When $\tau$ approximates to zero, the Gumbel-softmax operation approximately outputs a one-hot vector. With this softmax function, $p_{(i,j)}^k$ is the probability of selecting the method $k$ to model the feature interaction between feature $i$ and $j$. 
The candidate embeddings for a feature interaction are $\{e^{f}_{(i,j)}, e^{m}_{(i,j)}, e^{n}\}$ ($\{f,m,n\}$ indicates \emph{factorize}, \emph{memorize} and \emph{naïve} methods respectively). The output of combination module is formalized as the weighted sum over all the candidate embeddings of the current feature interaction 

\begin{equation}
\begin{split}
    \mathbf{e}^{b}_{(i,j)} &= \sum_{k \in \mathcal{K}} p^k_{(i,j)} \cdot \mathbf{e}^k_{(i,j)}  \\
    &= p^{m}_{(i,j)} \cdot \mathbf{e}^{m}_{(i,j)} + p^{f}_{(i,j)} \cdot \mathbf{e}^{f}_{(i,j)} +  p^{n}_{(i,j)} \cdot \mathbf{e}^{n}.
\end{split}
\end{equation}

Then the feature interaction embeddings $\mathbf{e}^{b}_{(i,j)}$ are fed into the MLP classifier so that all these parameters can be optimized via gradient descent. To summarize, with the Gumbel-softmax tricks, the search procedure becomes end-to-end differentiable.

To make the presentation more clear, we summarize the pseudo-code of the search algorithm in Algorithm \ref{alg:search}. The parameters that need to be optimized during the search period are in two categories: (\romannumeral1) $\Theta$, the model parameters of \textit{OptInter}, including the parameters of both Embedding tables and the MLP classifier; (\romannumeral2) $\alpha$, the architecture parameters selecting the optimal feature interaction methods from the given search space.

Following previous research work\cite{AutoFis}, we update both model parameters $\Theta$ and architecture parameters $\alpha$ simultaneously instead of alternately. This is because, in CTR prediction, the predicted label $\hat{y}$ is highly sensitive towards the embedding table. Suppose the model parameters $\Theta$ and architecture parameters $\alpha$ are trained alternately. In that case, the overall framework is hard to converge (and therefore resulted in suboptimal performance) because a small disturb in $\alpha$ leads to a significant change in $\Theta$. 
Moreover, we empirically compare the result of updating both model parameter $\Theta$ and architecture parameter $\alpha$ simultaneously and alternately in Section~\ref{sec:different_optim}, which demonstrates that our learning algorithm is more effective.

\subsubsection{Re-train}

In the search stage, architecture parameters also influence the model training. Re-training model with fixed architecture parameters can eliminate such influences of suboptimal modelling methods during the search stage. Hence, we introduce the re-train stage to fully train the model with the optimal method for each feature interaction that is found in the search stage. 

\begin{algorithm}
    \caption{The Optimization of Re-train Process}
    \label{alg:retrain}
    \begin{algorithmic}[1]
    	\Require Training dataset $\mathcal{D}$ and searched optimal architecture parameter $\alpha^{*}$
    	\Ensure the well-trained model parameter $\Theta^*$
        \While {not converge}
            \State Sample a mini-batch of training data $\mathcal{D}_{\text{train}}$
            \State Generate the predicted labels $\hat{y}$ via \textit{OptInter} with 
            \Statex \qquad current model parameter $\Theta$ and the optimal  
            \Statex \qquad architecture parameter $\alpha^{*}$ given Equation \ref{eq:y_hat}
            \State Calculate the cross-entropy loss $\mathcal{L}(\mathcal{D}_{\text{train}}|\Theta, \alpha^{*})$ over
            \Statex \qquad the mini-batch given Equation \ref{eq:loss}
            \State Update model parameter $\Theta$ by descending gradient 
            \Statex \qquad $\bigtriangledown_{\Theta} \mathcal{L}(\mathcal{D}_{\text{train}}|\Theta, \alpha^{*})$
        \EndWhile
    \end{algorithmic}
\end{algorithm}

\begin{table*}[!htbp]
    \renewcommand\arraystretch{1.00}
    \centering
    \caption{Dataset statistics}
    \begin{tabular}{c|ccccccc}
        \hline
        Dataset & \#samples & \#cont & \#cate & \#cross & \#orig value & \#cross value & pos ratio \\
        \hline
        Criteo  & $4.6 \times 10^7$ & 13 & 26 & 325 & $5.1 \times 10^5$ & $3.7 \times 10^7$ & 0.23 \\
        Avazu   & $4.0 \times 10^7$ & 0  & 24 & 276 & $1.2 \times 10^6$ & $2.4 \times 10^8$ & 0.17 \\
        iPinYou & $1.9 \times 10^7$ & 0  & 16 & 120 & $9.4 \times 10^5$ & $6.8 \times 10^7$ & 0.0008 \\
        Private & $8.0 \times 10^8$ & 0  & 9  & 36  & $4.0 \times 10^5$ & $7.1 \times 10^7$ & 0.17 \\
        \hline
    \end{tabular}
    \begin{tablenotes}
    \footnotesize
    \item[1] Note: \textit{\#cont} refers to the number of the continuous original feature, \textit{\#cate} refers to the number of the categorical original feature, \textit{\#cross} refers to the number of the cross-product transformed features, \textit{\#orig value} refers to the number of unique values for original features, \textit{\#cross value} refers to the number of unique values for cross-product transformed features, \textit{pos ratio} refers to the positive ratio. 
    \end{tablenotes}
    \label{Table:dataset}
\end{table*}

During the re-train stage, the gumbel-softmax operation is no longer used. We select the optimal modelling method for each feature interaction with the largest weight, based on the learned parameter $\alpha$. This is formalized as

\begin{equation}
\begin{split}
    \mathbf{e}^{b}_{(i,j)} &= \mathbf{e}_{(i,j)}^{k^{*}}, \\
    \text{s.t. $k^{*}$} &= \argmax_{k \in \mathcal{K}} \alpha_{(i,j)}^k.
\end{split}
\end{equation}

In all, we re-train the model following Algorithm \ref{alg:retrain} after obtaining the optimal modelling method for each feature interaction. 

\subsection{Model Discussion}
\label{model}

\begin{table}[!htbp]
    \renewcommand\arraystretch{1.00}
    \centering
    \caption{Model Discussion}
    \begin{tabular}{c|c|c|c|c}
    \hline
        \multirow{2}{*}{Category} & \multirow{2}{*}{Model} & \multicolumn{2}{|c|}{Feature Interaction Layer} & \multirow{2}{*}{Classifier} \\
        \cline{3-4}
        & & Method & Func. \\
    \hline
        \multirow{2}{*}{\emph{naïve}} 
        & LR\cite{LR} & $\{n\}$ & - & Shallow \\
        & FNN\cite{FNN} & $\{n\}$ & - & Deep \\
    \hline
        \multirow{2}{*}{\emph{memorized}} 
        & Poly2\cite{Poly-2} & $\{m\}$  & - & Shallow \\
        & Wide\&Deep\cite{Wide_Deep} & $\{m\}$ & - & S\&D \\
    \hline
        \multirow{8}{*}{\emph{factorized}}
        & FM\cite{FM} & $\{f\}$ & $\langle \mathbf{e}^o_i, \mathbf{e}^o_j \rangle$ & Shallow \\
        & FwFM\cite{FwFM} & $\{f\}$ & $\langle \mathbf{e}^o_i, \mathbf{e}^o_j \rangle w_{(i,j)}$ & Shallow \\
        & FmFM\cite{FM2} & $\{f\}$ & $ \mathbf{e}^o_i W_{(i,j)} {\mathbf{e}^o_j}^T $ & Shallow \\
        & IPNN\cite{PNN16} & $\{f\}$ & $\langle \mathbf{e}^o_i, \mathbf{e}^o_j \rangle$ & Deep \\
        & OPNN\cite{PNN16} & $\{f\}$ & $\langle \mathbf{e}^o_i, \mathbf{e}^o_j \rangle_{\phi}$ & Deep \\
        & DeepFM\cite{DeepFM} & $\{f\}$ & $\langle \mathbf{e}^o_i, \mathbf{e}^o_j \rangle$ & Deep \\
        & PIN\cite{PNN19} & $\{f\}$ & net($\mathbf{e}^o_i, \mathbf{e}^o_j$) & Deep \\
    \hline
        \multirow{2}{*}{\emph{hybrid}}
        & AutoFIS\cite{AutoFis} & $\{n,f\}$ & flexible & Deep \\
        & \textit{OptInter} & $\{n,m,f\}$ & flexible & Deep \\
    \hline
    \end{tabular}
    \begin{tablenotes}
    \footnotesize
    \item[1] Note: \textit{Category} refers to which category does the model belong to, which is determined by how it model feature interaction. All models can generally be categorized into four classes: \emph{naïve}, \emph{memorized}, \emph{factorized} and \emph{hybrid}. \textit{Method} denotes the potential methods, with $n$, $m$, $f$ denoting \emph{naïve}, \emph{memorized}, \emph{factorized} methods respectively. The \textit{Method} is fixed unless the model belongs to \emph{Hybrid} category. \textit{Func.} refers to the factorization function, which is only meaningful for factorized method. \textit{net} refers to a neural network. \textit{S\&D} is short for shallow and deep.
    \end{tablenotes}
    \label{table:model_discussion}
\end{table}

In this section, we discuss the relationship of \textit{OptInter} with mainstream CTR models. We distinguish these models by the method and factorization function in feature interaction layer. As summarized in Table \ref{table:model_discussion}, all the models can be viewed as instances of our framework. Furthermore, some detailed conclusions can be observed as follows:
\begin{itemize}
    \item According to the modelling methods they use in the feature interaction layer, these methods are grouped into four categorizes: \emph{naïve} methods~\cite{LR,FNN}; \emph{memorized} methods~\cite{Poly-2,Wide_Deep}; \emph{factorized} methods~\cite{FM,FwFM,FM2,PNN16,PNN19,DeepFM}; and \emph{hybrid} methods~\cite{AutoFis} (including \textit{OptInter}).
     \item All the models can be viewd as instances of our \textit{OptInter} framework. For instance, FNN~\cite{FNN} is a deep model via the \emph{naïve} feature interaction method. IPNN~\cite{PNN16} \emph{factorizes} feature interaction with inner-product function.
    \item Most of the deep CTR models adopt \emph{factorized} methods, which only differ in the factorization functions. Our \textit{OptInter} is the first to introduce \emph{memorized} method in search space for deep CTR models. We empirically demonstrate the benefits of introducing \emph{memorized} method in deep CTR models.
    \item AutoFIS~\cite{AutoFis} searches suitable feature interactions to be \emph{factorized}, therefore it utilizes a hybrid modelling method over \{\emph{factorized}, \emph{naïve}\}. 
    The search space of \textit{OptInter} is a superset of AutoFIS. Note that both \textit{OptInter} and AutoFIS can be flexibly adopted to any factorzation function.
\end{itemize}

\section{Experiments}

In this section, to comprehensively evaluate our proposed framework, we design experiments to answer the following research questions: 

\begin{itemize}
    \item \textbf{RQ1}: Could \textit{OptInter} achieve superior performance, compared with mainstream CTR prediction models?
    \item \textbf{RQ2}: How effective is the \emph{memorized} method in deep CTR models?
    \item \textbf{RQ3}: Is the good performance achieved by \textit{OptInter} due to the increased amount of parameters?
    \item \textbf{RQ4}: How effective is two-stage learning algorithm in \textit{OptInter} (namely, search and re-train)?
    \item \textbf{RQ5}: What kind of feature interactions is selected by each modeling method in \textit{OptInter}?
\end{itemize}

\begin{table}[!htbp]
    \renewcommand\arraystretch{1.00}
	\centering
	\caption{Parameter Setup}
	\begin{tabular}{|c|c|c|c|}
		\hline
		    Params & Criteo & Avazu & iPinYou \\
		\hline
		    \multirow{5}{*}{General} & bs=2000 & bs=2000 & bs=2000 \\
		        & opt=Adam & opt=Adam & opt=Adam \\
		        & lr=5e-4 & lr=5e-4 & lr\_o=1e-5 \\
		        & l2\_o=0.0 & l2\_o=0.0 & l2\_o=1e-6 \\
		        & eps=1e-8 & eps=1e-8 & eps=1e-4 \\
		\hline
		    LR & - & - & lr\_o=1e-4 \\
		\hline
		    \multirow{4}{*}{FM} 
		    & \multirow{4}{*}{s1=20} & \multirow{4}{*}{s1=40} & s1=20 \\
		    & & & lr\_o=1e-4 \\
		    & & & l2\_o=1e-9 \\
		    & & & eps=1e-6 \\
		\hline
		    Poly-2 & - & - & - \\
		\hline
		    \multirow{2}{*}{FNN} 
		    & s1=20 & s1=40 & s1=20 \\
		    \multirow{2}{*}{IPNN} 
		    & net=[$700 \times 5$] & net=[$500 \times 5$] & net=[$300 \times 3$] \\
		    & LN=true & LN=true & LN=true \\
		\hline
		    \multirow{4}{*}{DeepFM} 
		    & s1=20 & s1=40 & s1=20 \\
		    & lr=5e-4 & lr=5e-4 & l2\_o=1e-7 \\
		    & net=[$700 \times 5$] & net=[$500 \times 5$] & net=[$300 \times 3$] \\
		    & LN=true & LN=true & LN=true \\
		\hline
		    \multirow{5}{*}{PIN} & \multirow{2}{*}{s1=20} & \multirow{2}{*}{s1=40} & s1=20 \\
		        & \multirow{2}{*}{net=[$700 \times 5$]} & \multirow{2}{*}{net=[$500 \times 5$]} & net=[$300 \times 3$] \\
		        & \multirow{2}{*}{sub-net=[40,5]} & \multirow{2}{*}{sub-net=[40,5]} & sub-net=[40,5] \\
		        & \multirow{2}{*}{LN=true} & \multirow{2}{*}{LN=true} & LN=true  \\
		        & & & l2\_o=1e-9 \\
		\hline
		    \multirow{1}{*}{AutoFIS}  & mu=0.8, c=5e-4 & mu=0.8, c=5e-4 & mu=0.535, c=5e-3 \\
		\hline
		    \multirow{4}{*}{\textit{OptInter-M}} & s1=20, s2=10 & s1=40, s2=4 & s1=20, s2=2 \\
		    \multirow{4}{*}{\textit{OptInter-F}} & lr\_o=lr\_c=1e-4 & lr\_o=lr\_c=1e-4 & lr\_c=1e-6 \\
		        & l2\_c=3e-8 & l2\_c=3e-8 & l2\_c=3e-8 \\
		        & net=[$700 \times 5$] & net=[$500 \times 5$] & net=[$300 \times 3$] \\
		        & LN=true & LN=true & LN=true \\
		    \textit{OptInter} & lr\_a=3e-5 & lr\_a=3e-5 & lr\_a=1e-3 \\
		\hline
	\end{tabular}
	\begin{tablenotes}
    \footnotesize
    \item[1] Note: bs=batch size, opt=optimizer, lr\_o=learning rate for original feature embedding table and neural network parameter, lr\_c=learning rate for feature combination embedding table, lr\_a=learning rate for architecture parameters,  l2\_o=l\_2 regularization on original feature embedding table, l2\_c=l\_2 regularization on feature combination embedding table, s1=embedding size for original feature, s2=embedding size for cross-product transformed feature, net=MLP structure, LN=layer normalization, mu and c are parameters in GRDA optimizer\cite{GRDA}.
    \end{tablenotes}
	\label{Table:param}
\end{table}

\subsection{Experiment Setup}
\subsubsection{Datasets}
We conduct our experiments on three public datasets and one private dataset. The statistics of all datasets are given in Table~\ref{Table:dataset}. We describe all these datasets and the pre-processing steps below.

\textbf{Criteo}\footnote{https://labs.criteo.com/2013/12/download-terabyte-click-logs/} dataset was used in a competition on click-through rate prediction jointly hosted by Criteo and Kaggle in 2014. 80\% of the randomly shuffled data is used for training and validation, with 20\% for testing. Both categorical features and cross-product transformed features with less than 20 times of appearance are set as a dummy feature out-of-vocabulary (OOV). Following~\cite{DeepFM}, the continuous features are first normalized into $[0,1]$ via min-max normalization technique
\begin{equation}
    x \gets \frac{x - x_{\min}}{x_{\max} - x_{\min}}
\end{equation}
and then multiplied with corresponding embeddings. 

\textbf{Avazu}\footnote{http://www.kaggle.com/c/avazu-ctr-prediction} dataset was released as a part of a click-through rate prediction challenge jointly hosted by Avazu and Kaggle in 2014. 80\% of the randomly shuffled data is used for training and validation, with 20\% for testing. Categorical features with less than five times of appearance are replaced by OOV.

\textbf{iPinYou}\footnote{https://contest.ipinyou.com/} dataset was released as a part of the RTB Bidding Algorithm Competition, 2013. We only use the click data from seasons 2 and 3 because of the same data schema. We follow the previous data processing \cite{PNN19}\footnote{https://github.com/wnzhang/make-ipinyou-data} and remove “user tags” to prevent leakage.

\textbf{Private} dataset is collected from Huawei App Store, which samples from user behaviour logs in eight consecutive days. We select the first seven days as training and validation set and the last day as testing set. This dataset contains app features (e.g., App ID, category), user features (e.g., user's behaviour history) and context features. 

\subsubsection{Metrics}

Following the previous works~\cite{DeepFM,PNN16,PNN19}, we use the common evaluation metrics for CTR prediction, \textbf{AUC} (Area Under ROC) and \textbf{Log loss} (cross-entropy). To measure the model size, we take the number of \textbf{parameters} in the model as the metric. Note that in CTR prediction task, $\mathbf{0.1 \%}$ improvement in terms of AUC is considered significant~\cite{Wide_Deep,FM2}.

\subsubsection{Baseline Models}

We select the most representative and state-of-the-art methods as our baselines: \change{\textit{LR}~\cite{LR} (logistic regression), \textit{FM}~\cite{FM} (factorization machine), \textit{Poly-2}~\cite{Poly-2} (logistic regression with all second-order feature interaction), \textit{FNN}~\cite{FNN} (deep neural network), \textit{IPNN}~\cite{PNN16} (deep neural network using inner-product as factorization function), \textit{DeepFM}~\cite{DeepFM} (deep neural network combined with factorization machine), \textit{PIN}~\cite{PNN19} (deep neural network using sub-neural network as factorization function) and \textit{AutoFIS}~\cite{AutoFis} (deep neural network with automatically selected second-order feature interaction). As discussed in Section \ref{model}, all these models can be viewed as instances of \textit{OptInter}.}

\info{R1O2: Here we introduce the baseline models following reviewers' suggestion.}

We also compare \change{another} two instances of our \textit{OptInter}: \textit{OptInter-M} and \textit{OptInter-F}. \textit{OptInter-F} only models feature interaction by factorizing latent vectors through the Hadamard product. \textit{OptInter-M} models feature interactions only by memorizing them as new features. 

\subsubsection{Parameter Setup}

To enable other researchers to reproduce our experiment results, we summarize all the hyper-parameters for each model in Table \ref{Table:param}. 
We will make the source code publicly available upon the acceptance of this paper.

For Criteo, Avazu and iPinYou datasets, the parameters of baseline models are set following~\cite{PNN19}. For \textit{OptInter-F} and \textit{OptInter-M} models, we grid search for the optimal hyperparameters. Following~\cite{PNN19}, Adam optimizer and Xavier initialization~\cite{Xavier} are adopted. Xavier initialises the weights in the model such that their values are subjected to a uniform distribution between $[-\sqrt{6/(n_{in}+n_{out})}$, $ \sqrt{6/(n_{in}+n_{out})}]$ with $n_{in}$ and $n_{out}$ being the input and output sizes of a hidden layer. Such initialization has been proven to be able to stabilize activations and gradients in the early stage of training~\cite{PNN19}. Layer normalization~\cite{LayerNorm} has been applied to each fully connected layer to avoid the internal covariate shift.

\subsubsection{Significance Test} \info{R2O1: We repeat the best baseline models and \textit{OptInter} 10 times.}\change{Following previous works\cite{AutoFis,AutoFeature},} 
we calculate the p-values for \textit{OptInter} and the best baseline by repeating the experiments \change{ten} times and performing a two-tailed pairwise t-test.

\subsubsection{Platform}
All experiments are conducted on a Linux server with 18 Intel Xeon Gold 6154 cores, 128 GB memory and one Nvidia-V100 GPU with PCIe connections.

\subsection{Overall Performance (RQ1 \& RQ2)}
\label{overall}

\info{R2O1: We re-run all models ten times and report the average result.}

\begin{table*}[!htbp]
    \renewcommand\arraystretch{1.00}
	\centering
	\change{
	\caption{Overall Performance Comparison}	\label{Table:overall}
	\begin{tabular}{c|ccc|ccc|ccc|ccc}
		\hline
		  \multicolumn{1}{c|}{Dataset} & \multicolumn{3}{c|}{Criteo} & \multicolumn{3}{c|}{Avazu} & \multicolumn{3}{c|}{iPinYou} & \multicolumn{3}{c}{Private} \\
		\hline
		  \multicolumn{1}{c|}{Metric} & AUC & Log loss & Param. & AUC & Log loss & Param. & AUC & Log loss & Param. & AUC & log Loss & Param. \\
		\hline
		  LR      & 0.7785 & 0.4708 & 0.5M    & 0.7685 & 0.3862 & 1.2M    & 0.7554 & 0.005689 & 0.9M      & 0.7690 & 0.3836 & 0.4M \\
		  FNN     & 0.7995 & 0.4514 & 13M     & 0.7858 & 0.3761 & 51M     & 0.7780 & 0.005622 & 19M       & 0.8348 & 0.3353 & 32M  \\
		\hline
		  FM      & 0.7845 & 0.4681 & 10M     & 0.7826 & 0.3790 & 49M     & 0.7776 & 0.005573 & 19M       & 0.8304 & 0.3406 & 32M  \\
		  IPNN    & 0.8005 & 0.4504 & 13M     & 0.7885 & 0.3745 & 51M     & 0.7784 & 0.005628 & 19M       & 0.8410 & 0.3303 & 32M  \\
		  DeepFM  & 0.7997 & 0.4511 & 13M     & 0.7860 & 0.3760 & 51M     & 0.7791 & 0.005617 & 19M       & 0.8383 & 0.3325 & 32M  \\
		  PIN     & 0.8016 & 0.4510 & 17M     & 0.7826 & 0.3790 & 52M     & 0.7782 & 0.005624 & 20M       & 0.8331 & 0.3365 & 33M  \\
		  \emph{OptInter-F}   & 0.8003 & 0.4507 & 21M     & 0.7860 & 0.3761 & 56M     & 0.7762 & 0.005688 & 23M       & 0.8380 & 0.3325 & 37M  \\
		\hline
		  Poly2   & 0.7827 & 0.4751 & 22M     & 0.7860 & 0.3795 & 241M    & 0.7740 & 0.005578 & 69M       & 0.8307 & 0.3390 & 71M  \\
		  \emph{OptInter-M}   & 0.8094 & 0.4423 & 225M    & 0.8060 & 0.3638 & 1012M   & 0.7800 & 0.005640 & 296M      & 0.8415 & 0.3265 & 738M \\
		\hline
		  AutoFIS & 0.8014 & 0.4514 & 22M     & 0.7861 & 0.3758 & 51M     & 0.7790 & 0.005618 & 19M         & 0.8413 & 0.3299 & 32M  \\
		  \textit{OptInter} & $\textbf{0.8101}^{*}$ & $\textbf{0.4417}^{*}$ & 100M & $\textbf{0.8062}^{*}$ & $\textbf{0.3637}^{*}$ & 827M & $\textbf{0.7825}^{*}$ & $\textbf{0.005604}^{*}$ & 26M & $\textbf{0.8425}^{*}$ & $\textbf{0.3256}^{*}$ & 302M\\

		\hline
	\end{tabular}

	\begin{tablenotes}
    \footnotesize
    \item[1] \textit{Param.} represents the number of parameters for this model. $*$ denotes statistically significant improvement (measured by t-test with p-value $<0.005$) over the best baseline. We group all the models into four categories: \emph{naïve} methods, \emph{factorize} methods, \emph{memorize} methods and \emph{hybrid} methods.
    \end{tablenotes}
}
\end{table*}

The overall performance of our \textit{OptInter} and the baselines on four datasets are reported in Table \ref{Table:overall}. We also report the details of selected architectures in Table \ref{Table:arch}. Based on our previous taxonomy in Section \ref{model}, we group these models into four categories: \emph{naïve}, \emph{factorized}, \emph{memorized} and \emph{hybrid}. We can make the following observations from Table~\ref{Table:overall}.

First, our \textit{OptInter} is both effective and efficient by automatically searching the optimal feature interaction methods. On Criteo and Avazu dataset, it achieves better performance than the best baseline (\textit{OptInter-M}) but requires only approximately 50\% to 80\% of the model parameters. On iPinYou dataset, it significantly improves the model performance compared with the best baseline model \textit{OptInter-M} with less than 10\% of the model parameters. Finally, on the Private dataset, it significantly improves the model performance compared with the best baseline model \textit{OptInter-M} with around 40\% of the model parameters. 

Secondly, the \emph{memorized} method (namely, \textit{OptInter-M}) is effective in the deep CTR model. On all the four datasets, \textit{OptInter-M} is the best performed baseline.
This verifies the necessity of involving \emph{memorized} method in the search space of \textit{OptInter}.

Finally, we observe that simply \emph{memorized} all feature interaction results in a very large model. Particularly, the number of parameters for \textit{OptInter-M} increases $10\times \sim 20\times$ compared with other baseline models. Across different datasets, our \emph{OptInter} selects suitable and necessary feature interactions to be \emph{memorized}. 
According to Table \ref{Table:arch}, \emph{OptInter} selects only 36\%, 40\% and 20\% feature interactions to be \emph{memorized} on three public datasets respectively. \textit{OptInter} is hence $2\times$, $1.22\times$ and $11.11\times$ smaller than \textit{OptInter-M} in terms of model size. With smaller model size, the performance of \textit{OptInter} is still higher than (or at least comparable to) \textit{OptInter-M}, which demonstrates that \textit{OptInter} is able to find suitable and necessary feature interactions to be memorized. Careful readers may find that \textit{OptInter} memorizes 40\% of the feature interactions but reduces only 20\% model size compared to  \textit{OptInter-M}, on Avazu dataset. The reason is that the feature \textit{Device\_ID} on Avazu dataset includes many distinct feature values, such that the feature interactions involving \textit{Device\_ID} have significantly more unique values than other feature interactions.

\begin{table}[!htbp]
    \renewcommand\arraystretch{1.00}
	\centering
	\caption{Method selection for different feature interactions}
	\begin{tabular}{c|ccc}
		\hline
		  Method & Criteo & Avazu & iPinYou \\
		\hline
		  Naive & [0,0,325] & [0,0,276] & [0,0,120] \\
		  \textit{OptInter-M} & [325,0,0] & [276,0,0] & [120,0,0] \\
		  \textit{OptInter-F} & [0,325,0] & [0,276,0] & [0,120,0] \\ 
		  AutoFIS & [0,54,271] & [0,58,218] & [0,112,8]\\
		  \textit{OptInter} & [117,98,110] & [107,73,96] & [25,12,83] \\
		\hline
	\end{tabular}
	\begin{tablenotes}
    \footnotesize
    \item[1] \textit{[x,y,z]} indicates the number of feature interactions that are selected to perform \emph{memorize}, \emph{factorize} and \emph{naïve} methods. 
    \end{tablenotes}
	\label{Table:arch}
\end{table}

\subsection{Detailed comparison with naïve and factorized methods (RQ3)}
\label{sec:same_budget}

\begin{table*}[!htbp]
    \renewcommand\arraystretch{1.05}
	\centering
	\caption{Performance Comparison with \emph{naïve} and \emph{factorized} models utilizing the same amount of parameters}
	\begin{tabular}{c|cccc|c|cccc|c}
		\hline
		  Dataset & \multicolumn{5}{c|}{Criteo} & \multicolumn{5}{c}{Avazu} \\
		\hline
		  Model         & FM        & FNN       & IPNN      & DeepFM    & OptInter      & FNN      & FM        & IPNN      & DeepFM    & OptInter \\
		\hline
		  AUC           & 0.7543    & 0.7990    & 0.8014    & 0.7678    & $\textbf{0.8101}^{*}$     & 0.7677    & 0.7848    & 0.7923    & 0.7691    & $\textbf{0.8062}^{*}$ \\
		  log loss      & 0.5192    & 0.4516    & 0.4495    & 0.5075    & $\textbf{0.4417}^{*}$     & 0.3947    & 0.3768    & 0.3723    & 0.3934    & $\textbf{0.3637}^{*}$ \\
		  Orig.E.       & 200       & 200       & 200       & 200       & 20              & 700     & 700       & 700       & 700       & 40 \\
		  Cross.E.       & 0         & 0         & 0         & 0         & 10              & 0       & 0         & 0         & 0         & 4 \\
		  Param.        & 103M      & 109M      & 109M      & 109M      & 100M            & 860M    & 953M      & 954M      & 953M      & 827M \\
		\hline
	\end{tabular}
	\begin{tablenotes}
    \footnotesize
    \item[1] \textit{Orig.E.} stands for embedding size of original features. \textit{Cross.E.} stands for embedding size of cross-product transformed features. \textit{Param.} represents the number of model parameters. $*$ denotes statistically significant improvement (measured by t-test with p-value $<0.005$ over the best baseline. 
    \end{tablenotes}
	\label{Table:same_budget}
\end{table*}

From Table~\ref{Table:overall}, \textit{OptInter} achieves much better performance than \emph{naïve} and \emph{factorized} methods, \change{at the cost of more parameters} on Criteo and Avazu datasets.
In this section, we compare \textit{OptInter} with \emph{naïve} and \emph{factorized} models given roughly the same amount of model parameters. 

We conduct this comparison experiments on Criteo and Avazu datasets, as \textit{OptInter} maintains comparable number of parameters as \emph{naïve} and \emph{factorized} methods on iPinYou dataset. The number of parameters of baseline \emph{naïve} and \emph{factorized} methods is increases by enlarging the embedding size ($20\times$ and $17.5\times$ for Criteo and Avazu).

The experimental result of such comparison is reported in Table~\ref{Table:same_budget}. 
As can be observed, enlarging embedding size does not bring significant performance improvement for \emph{naïve} and \emph{factorized} methods. Moreover, the model performance is even worse with a larger embedding size (e.g., FM and DeepFM) due to overfitting.
With the same amount of parameters, \textit{OptInter} still significantly outperforms \emph{naïve} and \emph{factorized} methods. The phenomenon indicates that when more space resource is available, enlarging embedding size is not always a good way to utilize such extra space to improve model performance. Instead, the experimental result demonstrates that increasing the number of parameters by searching suitable and necessary feature interactions to memorize as few features (as \textit{OptInter} does) is more effective to boost performance when extra space resource is provided. 

\subsection{Detailed comparison with memorized methods (RQ3)}

\begin{figure}[!htbp]
\centering
\subfigure[Criteo]{
\begin{minipage}[t]{0.22\textwidth}
\centering
\includegraphics[width=\textwidth]{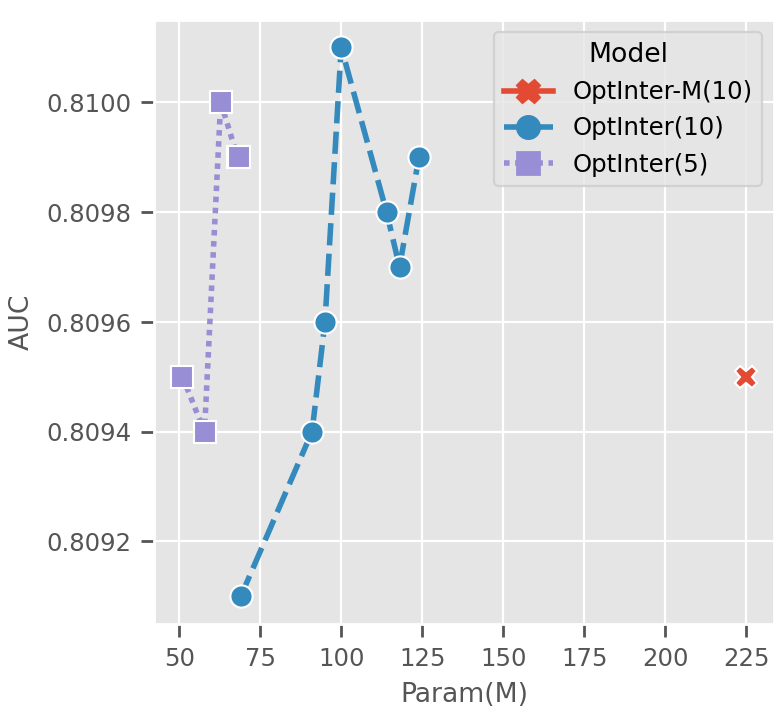}
\label{fig:param_AUC_Criteo}
\end{minipage}
}
\subfigure[Avazu]{
\begin{minipage}[t]{0.22\textwidth}
\centering
\includegraphics[width=\textwidth]{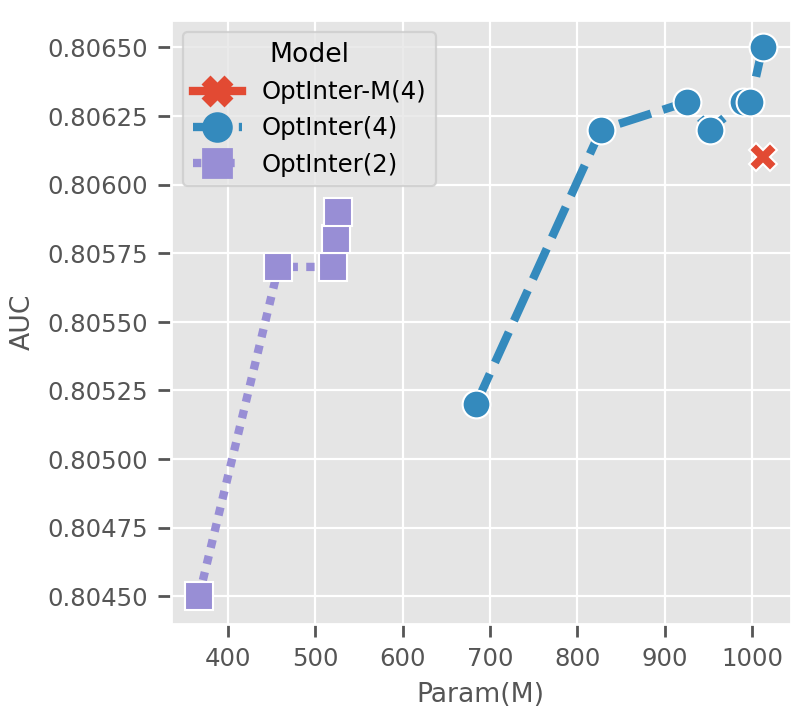}
\label{fig:param_AUC_Avazu}
\end{minipage}
}
\caption{Visualization of efficiency-effectiveness trade-off for different datasets. \change{Efficiency is measured by the size of parameters, shown in X-axis. Effectiveness is measured by the AUC score, shown in Y-axis.} Here \textit{OptInter-M(X)} indicates the embedding size of memorized embedding in \textit{OptInter-M} model being $X$, \textit{OptInter(Y)} indicates the embedding size of memorized embedding in \textit{OptInter} model being $Y$.
}
\label{fig:param_AUC}
\end{figure}

As demonstrated in Section \ref{overall}, \textit{OptInter-M} is the best baseline. 
In this section, we further compare \textit{OptInter} and \textit{OptInter-M} in terms of effectiveness (measured by AUC) and efficiency (measured by the number of parameters) on Criteo and Avazu datasets. 
As is shown in Figure \ref{fig:param_AUC}, three models are compared: \textit{OptInter-M}(10) sets the embedding size for memorized embeddings to 10, \textit{OptInter}(10) and \textit{OptInter}(5) sets the embedding size for memorized embeddings to 10 and 5. \info{R3O1: We add more detailed figure illustrations for Figure 4 and introduce the meaning of both the X-axis and Y-axis.} \change{In Figure \ref{fig:param_AUC_Criteo}, AUC varies from 0.8091 to 0.8101 and the number of parameters varies from 50M to 225M. In Figure \ref{fig:param_AUC_Avazu}, AUC varies from 0.8045 to 0.8065 and the number of parameters varies from 367M to 1012M.} 

From Figure~\ref{fig:param_AUC}, the following observations can be made.
First, \textit{OptInter} outperforms \textit{OptInter-M} under many cases with much fewer parameters. This is because \textit{OptInter} finds suitable and necessary feature interactions to memorize, instead of memorizing all of them as by \textit{OptInter-M}. 
Second, the curves of \textit{OptInter} indicate that its performance degrades dramatically when the number of parameters shrinks below a certain threshold. 
This reason is that some feature interactions are strong signals that must be \emph{memorized} to guarantee good performance (some such example feature interactions will be illustrated in Section \ref{sec:case_study}). 
Third, reducing the embedding size for cross-product transformed features can significantly decrease the number of parameters, with only a slight performance drop. This phenomenon suggests that when the space resource is constrained, reducing the embedding size of memorized embeddings is better than throwing away identified feature interactions to be memorized.

\subsection{Ablation study on the search stage (RQ4)}
\label{sec:different_optim}
In this section, we investigate how the search stage affects model performance on public datasets. We compare two search algorithms: \textit{Bi-level} and \textit{OptInter}. \textit{Bi-level} search algorithm updates model parameters and architecture parameters alternatively while the search algorithm in \textit{OptInter} optimizes these two families of parameters jointly. \textit{Bi-level} has been widely adopted in neural architecture search domain~\cite{DARTS}. However, as discussed in Section~\ref{sec:learning}, it may be suboptimal to update both model parameters and architecture parameters alternately as a minor updating in architecture parameters leads to a significant change in the model parameters (as also stated in~\cite{AutoFis}). 

\begin{table}[!htbp]
    \renewcommand\arraystretch{1.05}
	\centering
	\caption{Comparison between different search algorithm}
	\begin{tabular}{c|c|cccc}
		\hline
		  Dataset & Model & AUC & log loss & Arch & Param. \\
		\hline
		  \multirow{3}{*}{Criteo} 
          & Random    & 0.8089 & 0.3764 & -         & 84M \\
          & Bi-level  & 0.8099 & 0.3741 & [114,109,104] & 95M \\
          & OptInter  & \textbf{0.8101} & \textbf{0.3760} & [117,98,110] & 100M  \\
		\hline
		  \multirow{3}{*}{Avazu} 
		  & Random    & 0.8030 & 0.3658 & -         & 418M \\
		  & Bi-level  & \multicolumn{4}{c}{Out of Memory} \\
		  & OptInter  & \textbf{0.8062} & \textbf{0.3637} & [107,73,96] & 827M \\
		\hline
		  \multirow{3}{*}{iPinYou}
		  & Random    & 0.7781 & 0.005734 & [36,38,46] & 108M \\
		  & Bi-level  & 0.7796 & 0.005620 & [34,16,70] & 31M \\
		  & OptInter  & \textbf{0.7825} & \textbf{0.005606} & [25,12,83] & 26M \\
		\hline
	\end{tabular}
	\begin{tablenotes}
    \footnotesize
    \item[1] \textit{Arch} stands for the searched architecture. \textit{[x,y,z]} indicates the number of feature interactions that are selected to perform \emph{memorized}, \emph{factorized} and \emph{naïve} methods. \textit{Param.} represents the number of parameters for this model. \textit{Bi-level} refers to the result of architectures searched by bi-level optimization. For \textit{Random}, the mean performance of ten randomly generated architectures are reported here.
    \end{tablenotes}
	\label{Table:different_optim}
\end{table}

As a baseline, we also report the result of performing a random search (namely \textit{random}). With random search, we randomly assign each feature interaction with one of the three modelling methods. Then the mean values of all the metrics are reported by repeating the random search ten times. 

Observed from Table~\ref{Table:different_optim}, both two search algorithms, \textit{Bi-level} and \textit{OptInter}, outperform randomly generated architecture, which demonstrates the effectiveness of the search stage. The \textit{OptInter} performs better than \textit{Bi-level} method, which verifies our previous discussions on how on optimize model parameters and architecture parameters. 

Note that \emph{Bi-level} optimization requires approximate $\sim 2\times$ GPU memory compared with our search algorithm. Therefore, its experiment on the Avazu dataset cannot be accomplished due to the limits of GPU memory.

\subsection{Ablation study on the re-train stage (RQ4)}

In this section, we investigate how the re-train stage affects the model performance. 
We compare \textit{OptInter} performance under different settings with and without re-train stage, of which the result is presented in Table~\ref{Table:retrain}. 
It is observed that the re-train stage significantly improves the model performance. 
Without re-training, the candidate methods to model feature interactions may influence each other, which makes the neural network parameters $\Theta$ suboptimal before the learning process of architecture parameters converges.
Re-training makes neural network parameters optimal according to the suitable modelling method for each feature interaction decided in the search stage.

\begin{table}[!htbp]
    \renewcommand\arraystretch{1.05}
	\centering
	\caption{Performance Comparison between with or without re-train stage}
	\begin{tabular}{c|cc|cc}
		\hline
		  Dataset & \multicolumn{2}{c|}{Criteo} & \multicolumn{2}{c}{Avazu} \\
		\hline
		  Metric & w. & w.o. & w. & w.o. \\
		\hline
		  AUC         & 0.8101 & 0.7953 & 0.8062 & 0.7772 \\
		  log loss    & 0.3760 & 0.4558 & 0.3637 & 0.3829 \\
		\hline
	\end{tabular}
	\begin{tablenotes}
    \footnotesize
    \item[1] \textit{w.} stands for re-train stage after search stage with fixed architecture parameters. \textit{w.o.} stands for without re-train stage after search stage.
    \end{tablenotes}
	\label{Table:retrain}
\end{table}

These two sub-sections together demonstrate the effectiveness of the two-stage learning algorithm.

\subsection{Interpretable Discussions (RQ5)}

In this section, we investigate which kind of feature interactions are selected by \textit{OptInter} for each method (namely, \emph{memorize}, \emph{factorize} and \emph{naïve} methods) on Criteo and Avazu. 
We distinguish feature interactions with \emph{mutual information scores} between feature interactions and labels. For feature interactions $\{\mathcal{H}=(\mathbf{x}^o_i, \mathbf{x}^o_j)\}$ and ground truth labels $\mathbf{y}$ ($y \in \mathbf{y}$), the mutual information between them is defined as

\begin{equation}
\begin{split}
    \mathbf{MI}(\mathcal{\{H\}},\mathbf{y}) = &-\sum \mathbf{P}(y) \log \mathbf{P}(y) \\
    &+ \sum \mathbf{P}(\mathcal{H} ,y) \log \mathbf{P}(y|\mathcal{H}),
\end{split}
\end{equation}
where the first term is the marginal entropy and the second term is the conditional entropy of ground truth labels $\mathbf{y}$ given feature interaction $\mathcal{H} = (\mathbf{x}^o_{i},\mathbf{x}^o_{j})$. Note that feature interactions with high mutual information scores are more informative (hence more important) to the prediction. 

\subsubsection{Overall Comparison}

\begin{figure}[!htbp]
\centering
\subfigure[Criteo]{
\begin{minipage}[t]{0.225\textwidth}
\centering
\includegraphics[width=\textwidth]{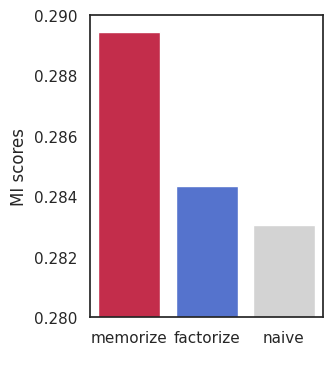}
\end{minipage}
}
\subfigure[Avazu]{
\begin{minipage}[t]{0.225\textwidth}
\centering
\includegraphics[width=\textwidth]{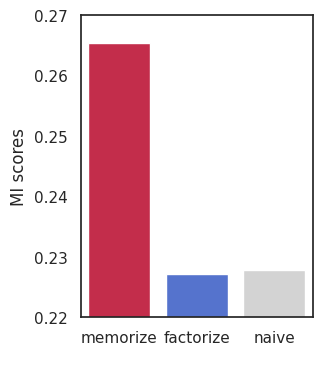}
\end{minipage}
}

\caption{Overall mutual information score of each method}
\label{fig:dataset} 
\end{figure}

We group feature interactions according to each individual method in \textit{OptInter}. We calculate and compare the mean mutual information scores of the feature interactions modelled by each method, as presented in Figure~\ref{fig:dataset}. Three conclusions can be observed. (\romannumeral1) \textit{OptInter} tends to memorize feature interactions with higher mutual information scores, which is reasonable because treating informative feature interactions as new features makes learning them better. (\romannumeral2) \textit{OptInter} removes the feature interactions with low mutual information scores, which helps remove uninformative and noisy feature interactions. (\romannumeral3) The characteristics of factorized feature interactions in \textit{OptInter} varies with respect to different datasets. 
To summarize, the overall trend of feature interactions selected by different methods is consistent with their mutual information scores, which explains the intuition of \textit{OptInter}. However, it is hard to assign a modelling method to each feature interaction based on heuristics as it may not be optimal, which motivates an automatic framework to do so.

\subsubsection{Case Study}
\label{sec:case_study}

\begin{figure*}[!htbp]
\centering
\subfigure[Heatmap of mutual information between each feature interaction and the label]{
\begin{minipage}[t]{0.35\textwidth}
\centering
\includegraphics[width=\textwidth]{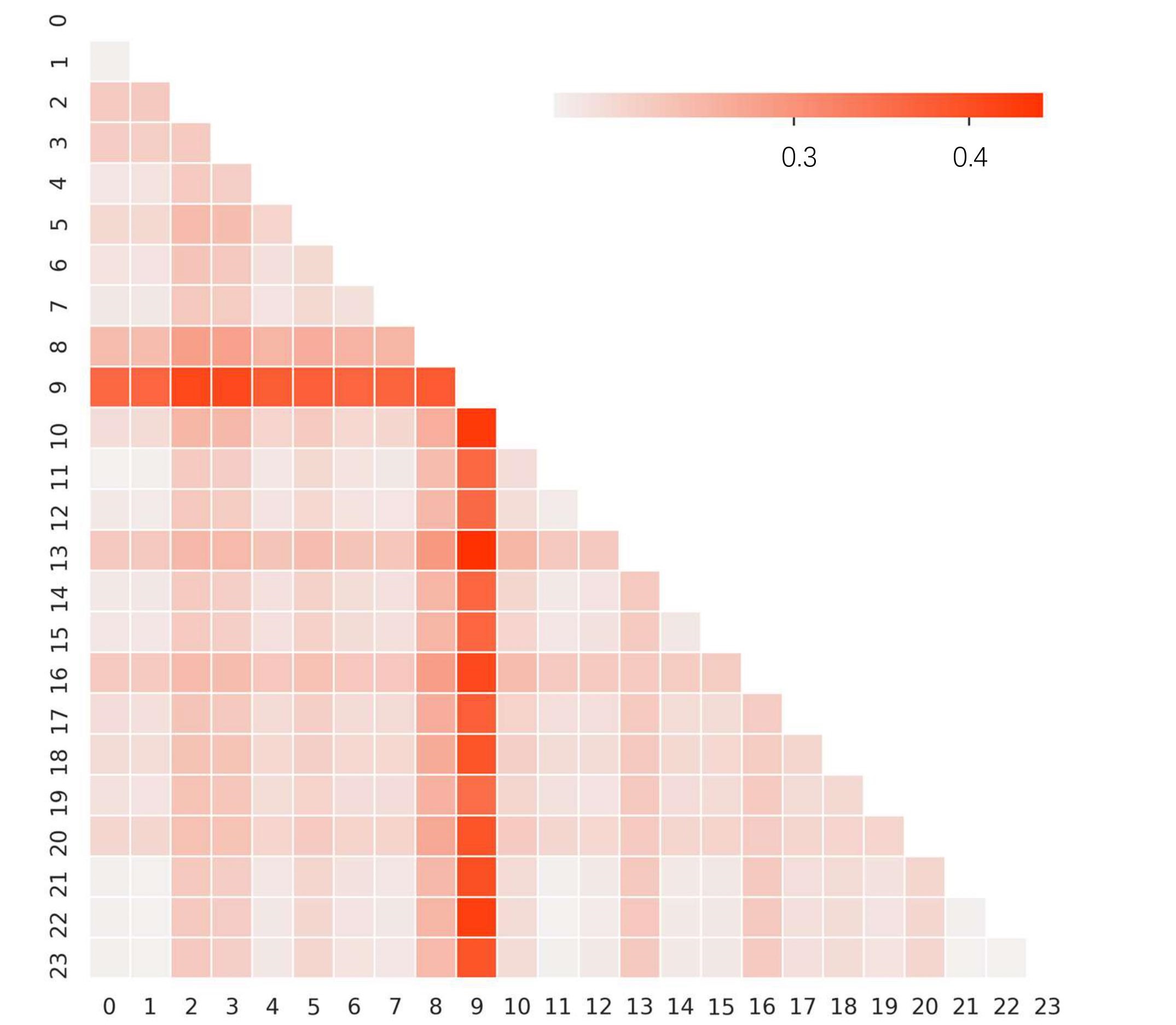}
\label{fig:avazu-MI}
\end{minipage}
}
\subfigure[Obtained optimal methods for feature interactions]{
\begin{minipage}[t]{0.35\textwidth}
\centering
\includegraphics[width=\textwidth]{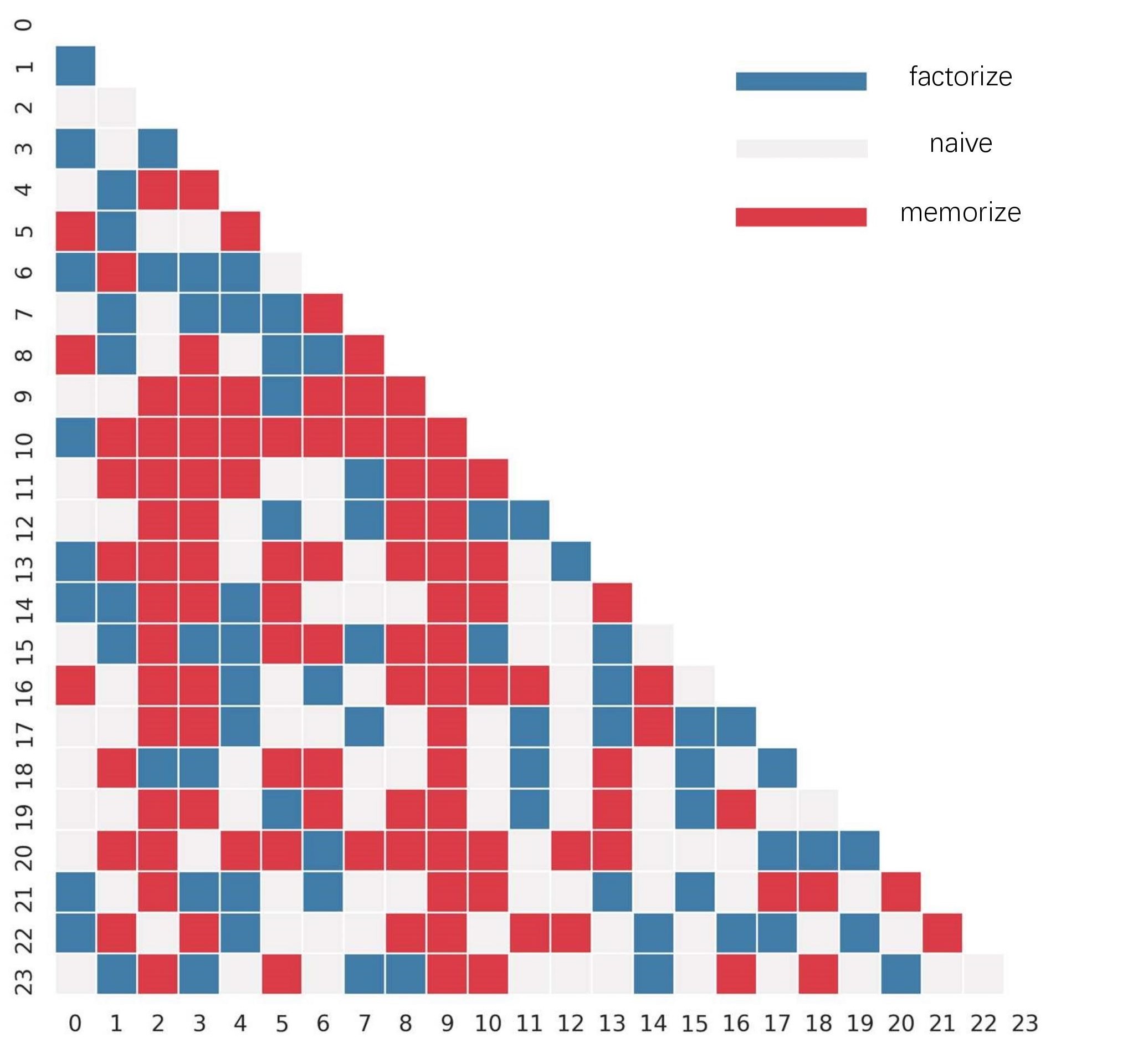}
\label{fig:avazu-Arch}
\end{minipage}
}
\caption{An example of interpretability on Avazu Dataset. \change{In both subfigures, the number indicates the field ID. In subfigure (a), the color indicates the strength of feature interaction in predicting label. In subfigure (b), different colors indicate three modelling methods searched by \textit{OptInter}.}}
\label{fig:avazu} 
\end{figure*}

As a case study, we investigate the selected method for each feature interaction \textit{OptInter} method on the Avazu dataset, which is shown in Figure~\ref{fig:avazu}. \info{R3O1: We add more detailed figure illustrations for Figure 6 and highlight the meaning of both subfigures.} Figure~\ref{fig:avazu-MI} shows the heat map of mutual information scores of all the feature interactions, which represents how informative each feature interaction is in predicting the label. Figure~\ref{fig:avazu-Arch} shows the searched method for each feature interaction. As can be seen, these two maps are positively correlated to each other, which indicates that \textit{OptInter} obtains the optimal method for each feature interaction.

\section{Related Work}
\label{sec:rw}

\subsection{Feature Interaction in CTR Prediction}

Feature interaction is one of the core problems in CTR prediction~\cite{FNN,PNN16,PNN19,DeepFM,AutoFeature,AutoFis,CAN}. Feature interaction is one of the core problems in CTR prediction~\cite{FNN,PNN16,PNN19,DeepFM,AutoFeature,AutoFis,CAN}. Generalized linear models like Logistic Regression (LR)~\cite{LR,GBDT} model feature interaction naively, i.e., they do not model feature interactions. One way to model all feature interactions is based on degree-2 polynomial mappings\cite{Poly-2}. It increases model performance, but it also leads to a combinatorial explosion. One possible solution is to design feature interactions manually. However, this requires experts' experience and is hard to generalize. Other methods like FM~\cite{FM} and its variants~\cite{FwFM,FFM} models the low order feature interactions by factorizing them with latent vectors.

With the success of deep learning in computer vision and natural language processes, the deep CTR prediction model has gained tremendous attention in recent years. Compared with the linear model, deep models can predict CTR more effectively. Usually, deep CTR models choose to model feature interaction into three classes: \emph{naïve}~\cite{FNN}, \emph{memorized}~\cite{Wide_Deep}, and \emph{factorized}~\cite{PNN16,PNN19,DeepFM}. FNN~\cite{FNN}, as an example of \emph{naïve} modelling, introduces multi-layer perceptron to predict CTR scores with original features. \change{However, such modelling method usually leads to lower model capacity (compared to the other two methods) because it cannot learn low-rank feature interactions well~\cite{Latent-Cross}. 

\info{R1O1: We move certain discussion from the introduction section into the related work section, and discuss the pros and cons of different modelling methods.}

\emph{Memorized} methods model feature interaction by memorizing them explicitly as new features. Wide\&Deep~\cite{Wide_Deep}, as an example, indicates that the \emph{memorized} method is a promising alternative for strong feature interactions. More specifically, Wide\&Deep~\cite{Wide_Deep} memorizes a manually selected set of feature interactions and feeds it into its wide component. In the experiments of~\cite{Wide_Deep}, the model structure for Google Play selects the feature interaction (\textit{User\_Installed\_App}, \textit{Impression\_App}) as new feature for its wide component. This is a powerful indicator for a user downloading an app, as the installation history indicates user preference. However, this requires experts' experience and is hard to generalize. Moreover, \emph{memorized} method induces a severe feature sparsity problem because the occurrence of many new features (generated by feature interactions) is relatively less than original features in the training set. The severe feature sparsity problem makes these infrequent new features difficult to learn and degrades the model performance. 

Due to this reason, \emph{factorized} methods are proposed to model feature interactions via a factorization function. Mainstream deep CTR models~\cite{DeepFM,PNN16,PNN19} usually adopt \emph{factorized} method to model feature interactions. For example, IPNN and OPNN~\cite{PNN16} use inner-product and outer-product as the factorization function, respectively. DeepFM~\cite{DeepFM} borrows the FM layer from Factorization Machine~\cite{FM} and uses that as a factorization function. PIN~\cite{PNN19}, improving on IPNN and OPNN~\cite{PNN16}, uses a small MLP component to model as a learnable factorization function. \emph{Factorized} methods alleviate feature sparsity issue as they assign learnable parameters to original features instead of assigning them to new features (generated by feature interactions) with a much larger feature space. However, these methods have their inherited drawbacks. As the feature interactions are modelled by the latent vectors of original features, the latent vectors take the responsibility of both representation learning and feature interaction modelling. These two tasks may conflict with each other so that the model performance is bounded, as observed in~\cite{ElementaryView,CAN}. 
}

Recently CAN~\cite{CAN} highlights the importance of \emph{memorized} method. It manually selects a subset of feature interactions to memorize and prove its effectiveness through extensive experiments. But their model design still makes CAN be a \emph{factorized} method.

\textit{OptInter} is the first work introducing the \emph{memorized} method in deep CTR models (although Wide\&Deep uses \emph{memorized} method, it applies \emph{memorized} method to its wide component only). Furthermore, \textit{OptInter} is general enough to unify mainstream deep CTR models under its framework.

\subsection{Neural architecture search (NAS) and its application in Recommender System}

Neural architecture search (NAS) aims at automatically finding an optimal architecture for specific tasks and datasets, which avoids manual design~\cite{MetaQNN,NAS,DARTS,NAO,Large-Scale-Evolution,Regularized-Evolution}. These methods could be used to find optimal network structures, loss functions or hyper-parameters, significantly reducing human intervention in the ML pipeline. These methods can be categorized into three classes: (\romannumeral1) reinforcement learning-based methods~\cite{MetaQNN,NAS} train an external controller (like RNN or reinforcement learning agent) to design cell structure for a specific model architecture; (\romannumeral2) evolutionary algorithms~\cite{Large-Scale-Evolution,Regularized-Evolution} that try to evolve an optimal architecture by mutating the top-$k$ architectures and explore new potential models; (\romannumeral3) gradient-based methods~\cite{NAO,DARTS} relax the search space to be continuous instead of searching on a discrete set of architectures. Such relaxation allows the searching process to be more efficient based on a gradient descent optimizer.

Recently, NAS methods have attracted much attention in the CTR prediction task. AutoFIS~\cite{AutoFis} utilizes GRDA Optimizer~\cite{GRDA} to select proper feature interactions. AutoFeature~\cite{AutoFeature} adopts a tree of Naive Bayes classifiers to find suitable feature interaction functions for various fields. AutoPI~\cite{AutoPI} extends the search space to computational graph and feature interaction functions to achieve higher generalization. Many research works~\cite{AutoDim,AutoEmb,ESAPN} aim to select suitable dimensions for various fields. AutoDis~\cite{AutoDis} focuses on modelling continuous feature and proposes a framework to select optimal discretization methods. AutoFT~\cite{AutoFT} proposes an end-to-end transfer learning framework to automatically determine transfer policy in CTR prediction via Gumbel-softmax tricks~\cite{Gumbel-Softmax}. AutoLoss~\cite{AutoLoss} proposes a framework that learn sample-specific loss function via bi-level optimization. 

\textit{OptInter} has its uniqueness compared with other research works that apply NAS techniques into the CTR prediction task. 
Compared with the works which focus on modelling feature interaction~\cite{AutoFis,AutoFeature,AutoPI}, \textit{OptInter} introduces \emph{memorized} methods into its search space and searches within a broader space than existing works.

\section{Conclusion}

In this paper, we proposed a novel deep CTR prediction framework named \textit{OptInter}, which is the first to introduce the \emph{memorized} feature interaction method in deep CTR models. \textit{OptInter} can search and identify the optimal method to model the feature interactions from \emph{naïve}, \emph{memorized} and \emph{factorized}. To achieve this, we first proposed a deep framework that unifies mainstream deep CTR models. Then, as a part of \textit{OptInter}, a two-stage learning algorithm is proposed. In the search stage, the search process is modelled as an architecture search problem solved efficiently by neural architecture search techniques. During the re-train stage, the model is re-trained from scratch to achieve better performance. Extensive experiments on four large-scale datasets demonstrate the superior performance of \textit{OptInter}. Several ablation studies show our method is effective in improving prediction performance and efficient in model size. Moreover, we also explain obtained results in the view of mutual information, which further highlights our method learns the optimal feature interaction methods.

\clearpage
\bibliographystyle{IEEEtran}
\bibliography{IEEEfull,ref.bib}

\begin{thebibliography}{10}
\providecommand{\url}[1]{#1}
\csname url@samestyle\endcsname
\providecommand{\newblock}{\relax}
\providecommand{\bibinfo}[2]{#2}
\providecommand{\BIBentrySTDinterwordspacing}{\spaceskip=0pt\relax}
\providecommand{\BIBentryALTinterwordstretchfactor}{4}
\providecommand{\BIBentryALTinterwordspacing}{\spaceskip=\fontdimen2\font plus
\BIBentryALTinterwordstretchfactor\fontdimen3\font minus
  \fontdimen4\font\relax}
\providecommand{\BIBforeignlanguage}[2]{{%
\expandafter\ifx\csname l@#1\endcsname\relax
\typeout{** WARNING: IEEEtran.bst: No hyphenation pattern has been}%
\typeout{** loaded for the language `#1'. Using the pattern for}%
\typeout{** the default language instead.}%
\else
\language=\csname l@#1\endcsname
\fi
#2}}
\providecommand{\BIBdecl}{\relax}
\BIBdecl

\bibitem{Wide_Deep}
H.~Cheng, L.~Koc, J.~Harmsen, T.~Shaked, T.~Chandra, H.~Aradhye, G.~Anderson,
  G.~Corrado, W.~Chai, M.~Ispir, R.~Anil, Z.~Haque, L.~Hong, V.~Jain, X.~Liu,
  and H.~Shah, ``Wide {\&} deep learning for recommender systems,'' in
  \emph{Proceedings of the 1st Workshop on Deep Learning for Recommender
  Systems, DLRS@RecSys 2016, Boston, MA, USA, September 15, 2016},
  A.~Karatzoglou, B.~Hidasi, D.~Tikk, O.~S. Shalom, H.~Roitman, B.~Shapira, and
  L.~Rokach, Eds.\hskip 1em plus 0.5em minus 0.4em\relax {ACM}, 2016, pp.
  7--10.

\bibitem{DeepFM}
H.~Guo, R.~Tang, Y.~Ye, Z.~Li, and X.~He, ``Deepfm: {A} factorization-machine
  based neural network for {CTR} prediction,'' in \emph{Proceedings of the
  Twenty-Sixth International Joint Conference on Artificial Intelligence,
  {IJCAI} 2017, Melbourne, Australia, August 19-25, 2017}, C.~Sierra, Ed.\hskip
  1em plus 0.5em minus 0.4em\relax ijcai.org, 2017, pp. 1725--1731.

\bibitem{PNN16}
Y.~Qu, H.~Cai, K.~Ren, W.~Zhang, Y.~Yu, Y.~Wen, and J.~Wang, ``Product-based
  neural networks for user response prediction,'' in \emph{{IEEE} 16th
  International Conference on Data Mining, {ICDM} 2016, December 12-15, 2016,
  Barcelona, Spain}, F.~Bonchi, J.~Domingo{-}Ferrer, R.~Baeza{-}Yates, Z.~Zhou,
  and X.~Wu, Eds.\hskip 1em plus 0.5em minus 0.4em\relax {IEEE} Computer
  Society, 2016, pp. 1149--1154.

\bibitem{PNN19}
Y.~Qu, B.~Fang, W.~Zhang, R.~Tang, M.~Niu, H.~Guo, Y.~Yu, and X.~He,
  ``Product-based neural networks for user response prediction over multi-field
  categorical data,'' \emph{{ACM} Trans. Inf. Syst.}, vol.~37, no.~1, pp.
  5:1--5:35, 2019.

\bibitem{FNN}
W.~Zhang, T.~Du, and J.~Wang, ``Deep learning over multi-field categorical data
  - - {A} case study on user response prediction,'' in \emph{Advances in
  Information Retrieval - 38th European Conference on {IR} Research, {ECIR}
  2016, Padua, Italy, March 20-23, 2016. Proceedings}, ser. Lecture Notes in
  Computer Science, N.~Ferro, F.~Crestani, M.~Moens, J.~Mothe, F.~Silvestri,
  G.~M.~D. Nunzio, C.~Hauff, and G.~Silvello, Eds., vol. 9626.\hskip 1em plus
  0.5em minus 0.4em\relax Springer, 2016, pp. 45--57.

\bibitem{Universal}
K.~Hornik, M.~B. Stinchcombe, and H.~White, ``Multilayer feedforward networks
  are universal approximators,'' \emph{Neural Networks}, vol.~2, no.~5, pp.
  359--366, 1989.

\bibitem{Latent-Cross}
A.~Beutel, P.~Covington, S.~Jain, C.~Xu, J.~Li, V.~Gatto, and E.~H. Chi,
  ``Latent cross: Making use of context in recurrent recommender systems,'' in
  \emph{Proceedings of the Eleventh {ACM} International Conference on Web
  Search and Data Mining, {WSDM} 2018, Marina Del Rey, CA, USA, February 5-9,
  2018}, Y.~Chang, C.~Zhai, Y.~Liu, and Y.~Maarek, Eds.\hskip 1em plus 0.5em
  minus 0.4em\relax {ACM}, 2018, pp. 46--54.

\bibitem{Poly-2}
Y.~Chang, C.~Hsieh, K.~Chang, M.~Ringgaard, and C.~Lin, ``Training and testing
  low-degree polynomial data mappings via linear {SVM},'' \emph{J. Mach. Learn.
  Res.}, vol.~11, pp. 1471--1490, 2010.

\bibitem{FM}
S.~Rendle, ``Factorization machines,'' in \emph{{ICDM} 2010, The 10th {IEEE}
  International Conference on Data Mining, Sydney, Australia, 14-17 December
  2010}, G.~I. Webb, B.~Liu, C.~Zhang, D.~Gunopulos, and X.~Wu, Eds.\hskip 1em
  plus 0.5em minus 0.4em\relax {IEEE} Computer Society, 2010, pp. 995--1000.

\bibitem{FFM}
Y.~Juan, Y.~Zhuang, W.~Chin, and C.~Lin, ``Field-aware factorization machines
  for {CTR} prediction,'' in \emph{Proceedings of the 10th {ACM} Conference on
  Recommender Systems, Boston, MA, USA, September 15-19, 2016}, S.~Sen,
  W.~Geyer, J.~Freyne, and P.~Castells, Eds.\hskip 1em plus 0.5em minus
  0.4em\relax {ACM}, 2016, pp. 43--50.

\bibitem{FwFM}
J.~Pan, J.~Xu, A.~L. Ruiz, W.~Zhao, S.~Pan, Y.~Sun, and Q.~Lu, ``Field-weighted
  factorization machines for click-through rate prediction in display
  advertising,'' in \emph{Proceedings of the 2018 World Wide Web Conference on
  World Wide Web, {WWW} 2018, Lyon, France, April 23-27, 2018}, P.~Champin,
  F.~Gandon, M.~Lalmas, and P.~G. Ipeirotis, Eds.\hskip 1em plus 0.5em minus
  0.4em\relax {ACM}, 2018, pp. 1349--1357.

\bibitem{FM2}
Y.~Sun, J.~Pan, A.~Zhang, and A.~Flores, ``Fm\({}^{\mbox{2}}\): Field-matrixed
  factorization machines for recommender systems,'' \emph{CoRR}, vol.
  abs/2102.12994, 2021.

\bibitem{ElementaryView}
S.~Prillo, ``An elementary view on factorization machines,'' in
  \emph{Proceedings of the Eleventh {ACM} Conference on Recommender Systems,
  RecSys 2017, Como, Italy, August 27-31, 2017}, P.~Cremonesi, F.~Ricci,
  S.~Berkovsky, and A.~Tuzhilin, Eds.\hskip 1em plus 0.5em minus 0.4em\relax
  {ACM}, 2017, pp. 179--183.

\bibitem{CAN}
G.~Zhou, W.~Bian, K.~Wu, L.~Ren, Q.~Pi, Y.~Zhang, C.~Xiao, X.~Sheng, N.~Mou,
  X.~Luo, C.~Zhang, X.~Qiao, S.~Xiang, K.~Gai, X.~Zhu, and J.~Xu, ``{CAN:}
  revisiting feature co-action for click-through rate prediction,''
  \emph{CoRR}, vol. abs/2011.05625, 2020.

\bibitem{AutoFis}
B.~Liu, C.~Zhu, G.~Li, W.~Zhang, J.~Lai, R.~Tang, X.~He, Z.~Li, and Y.~Yu,
  ``Autofis: Automatic feature interaction selection in factorization models
  for click-through rate prediction,'' in \emph{{KDD} '20: The 26th {ACM}
  {SIGKDD} Conference on Knowledge Discovery and Data Mining, Virtual Event,
  CA, USA, August 23-27, 2020}, R.~Gupta, Y.~Liu, J.~Tang, and B.~A. Prakash,
  Eds.\hskip 1em plus 0.5em minus 0.4em\relax {ACM}, 2020, pp. 2636--2645.

\bibitem{DARTS}
H.~Liu, K.~Simonyan, and Y.~Yang, ``{DARTS:} differentiable architecture
  search,'' \emph{CoRR}, vol. abs/1806.09055, 2018.

\bibitem{Large-Scale-Evolution}
E.~Real, S.~Moore, A.~Selle, S.~Saxena, Y.~L. Suematsu, J.~Tan, Q.~V. Le, and
  A.~Kurakin, ``Large-scale evolution of image classifiers,'' in
  \emph{Proceedings of the 34th International Conference on Machine Learning,
  {ICML} 2017, Sydney, NSW, Australia, 6-11 August 2017}, ser. Proceedings of
  Machine Learning Research, D.~Precup and Y.~W. Teh, Eds., vol.~70.\hskip 1em
  plus 0.5em minus 0.4em\relax {PMLR}, 2017, pp. 2902--2911.

\bibitem{AutoFeature}
F.~Khawar, X.~Hang, R.~Tang, B.~Liu, Z.~Li, and X.~He, ``Autofeature: Searching
  for feature interactions and their architectures for click-through rate
  prediction,'' in \emph{{CIKM} '20: The 29th {ACM} International Conference on
  Information and Knowledge Management, Virtual Event, Ireland, October 19-23,
  2020}, M.~d'Aquin, S.~Dietze, C.~Hauff, E.~Curry, and
  P.~Cudr{\'{e}}{-}Mauroux, Eds.\hskip 1em plus 0.5em minus 0.4em\relax {ACM},
  2020, pp. 625--634.

\bibitem{Gumbel-Softmax}
E.~Jang, S.~Gu, and B.~Poole, ``Categorical reparameterization with
  gumbel-softmax,'' in \emph{5th International Conference on Learning
  Representations, {ICLR} 2017, Toulon, France, April 24-26, 2017, Conference
  Track Proceedings}.\hskip 1em plus 0.5em minus 0.4em\relax OpenReview.net,
  2017.

\bibitem{AutoPI}
Z.~Meng, J.~Zhang, Y.~Li, J.~Li, T.~Zhu, and L.~Sun, ``A general method for
  automatic discovery of powerful interactions in click-through rate
  prediction,'' \emph{CoRR}, vol. abs/2105.10484, 2021.

\bibitem{FIVES}
Y.~Xie, Z.~Wang, Y.~Li, B.~Ding, N.~M. G{\"{u}}rel, C.~Zhang, M.~Huang, W.~Lin,
  and J.~Zhou, ``Interactive feature generation via learning adjacency tensor
  of feature graph,'' \emph{CoRR}, vol. abs/2007.14573, 2020.

\bibitem{LayerNorm}
L.~J. Ba, J.~R. Kiros, and G.~E. Hinton, ``Layer normalization,'' \emph{CoRR},
  vol. abs/1607.06450, 2016.

\bibitem{Gumbel-Softmax-dist}
E.~J. Gumbel, ``Categorical reparameterization with gumbel-softmax,'' in
  \emph{Statistical theory of extreme values and some practical applications: a
  series of lectures, Vol. 33, US}.\hskip 1em plus 0.5em minus 0.4em\relax US
  Government Printing Office, 1945.

\bibitem{LR}
M.~Richardson, E.~Dominowska, and R.~Ragno, ``Predicting clicks: estimating the
  click-through rate for new ads,'' in \emph{Proceedings of the 16th
  International Conference on World Wide Web, {WWW} 2007, Banff, Alberta,
  Canada, May 8-12, 2007}, C.~L. Williamson, M.~E. Zurko, P.~F.
  Patel{-}Schneider, and P.~J. Shenoy, Eds.\hskip 1em plus 0.5em minus
  0.4em\relax {ACM}, 2007, pp. 521--530.

\bibitem{GRDA}
S.~Chao, Z.~Wang, Y.~Xing, and G.~Cheng, ``Directional pruning of deep neural
  networks,'' in \emph{Advances in Neural Information Processing Systems 33:
  Annual Conference on Neural Information Processing Systems 2020, NeurIPS
  2020, December 6-12, 2020, virtual}, H.~Larochelle, M.~Ranzato, R.~Hadsell,
  M.~Balcan, and H.~Lin, Eds., 2020.

\bibitem{Xavier}
X.~Glorot and Y.~Bengio, ``Understanding the difficulty of training deep
  feedforward neural networks,'' in \emph{Proceedings of the Thirteenth
  International Conference on Artificial Intelligence and Statistics, {AISTATS}
  2010, Chia Laguna Resort, Sardinia, Italy, May 13-15, 2010}, ser. {JMLR}
  Proceedings, Y.~W. Teh and D.~M. Titterington, Eds., vol.~9.\hskip 1em plus
  0.5em minus 0.4em\relax JMLR.org, 2010, pp. 249--256.

\bibitem{GBDT}
X.~He, J.~Pan, O.~Jin, T.~Xu, B.~Liu, T.~Xu, Y.~Shi, A.~Atallah, R.~Herbrich,
  S.~Bowers, and J.~Q. Candela, ``Practical lessons from predicting clicks on
  ads at facebook,'' in \emph{Proceedings of the Eighth International Workshop
  on Data Mining for Online Advertising, {ADKDD} 2014, August 24, 2014, New
  York City, New York, {USA}}, E.~Saka, D.~Shen, K.~Lee, and Y.~Li, Eds.\hskip
  1em plus 0.5em minus 0.4em\relax {ACM}, 2014, pp. 5:1--5:9.

\bibitem{MetaQNN}
B.~Baker, O.~Gupta, N.~Naik, and R.~Raskar, ``Designing neural network
  architectures using reinforcement learning,'' in \emph{5th International
  Conference on Learning Representations, {ICLR} 2017, Toulon, France, April
  24-26, 2017, Conference Track Proceedings}.\hskip 1em plus 0.5em minus
  0.4em\relax OpenReview.net, 2017.

\bibitem{NAS}
B.~Zoph and Q.~V. Le, ``Neural architecture search with reinforcement
  learning,'' in \emph{5th International Conference on Learning
  Representations, {ICLR} 2017, Toulon, France, April 24-26, 2017, Conference
  Track Proceedings}.\hskip 1em plus 0.5em minus 0.4em\relax OpenReview.net,
  2017.

\bibitem{NAO}
R.~Luo, F.~Tian, T.~Qin, E.~Chen, and T.~Liu, ``Neural architecture
  optimization,'' in \emph{Advances in Neural Information Processing Systems
  31: Annual Conference on Neural Information Processing Systems 2018, NeurIPS
  2018, December 3-8, 2018, Montr{\'{e}}al, Canada}, S.~Bengio, H.~M. Wallach,
  H.~Larochelle, K.~Grauman, N.~Cesa{-}Bianchi, and R.~Garnett, Eds., 2018, pp.
  7827--7838.

\bibitem{Regularized-Evolution}
E.~Real, A.~Aggarwal, Y.~Huang, and Q.~V. Le, ``Regularized evolution for image
  classifier architecture search,'' in \emph{The Thirty-Third {AAAI} Conference
  on Artificial Intelligence, {AAAI} 2019, The Thirty-First Innovative
  Applications of Artificial Intelligence Conference, {IAAI} 2019, The Ninth
  {AAAI} Symposium on Educational Advances in Artificial Intelligence, {EAAI}
  2019, Honolulu, Hawaii, USA, January 27 - February 1, 2019}.\hskip 1em plus
  0.5em minus 0.4em\relax {AAAI} Press, 2019, pp. 4780--4789.

\bibitem{AutoDim}
X.~Zhao, H.~Liu, H.~Liu, J.~Tang, W.~Guo, J.~Shi, S.~Wang, H.~Gao, and B.~Long,
  ``Memory-efficient embedding for recommendations,'' \emph{CoRR}, vol.
  abs/2006.14827, 2020.

\bibitem{AutoEmb}
X.~Zhao, C.~Wang, M.~Chen, X.~Zheng, X.~Liu, and J.~Tang, ``Autoemb: Automated
  embedding dimensionality search in streaming recommendations,'' \emph{CoRR},
  vol. abs/2002.11252, 2020.

\bibitem{ESAPN}
H.~Liu, X.~Zhao, C.~Wang, X.~Liu, and J.~Tang, ``Automated embedding size
  search in deep recommender systems,'' in \emph{Proceedings of the 43rd
  International {ACM} {SIGIR} conference on research and development in
  Information Retrieval, {SIGIR} 2020, Virtual Event, China, July 25-30, 2020},
  J.~Huang, Y.~Chang, X.~Cheng, J.~Kamps, V.~Murdock, J.~Wen, and Y.~Liu,
  Eds.\hskip 1em plus 0.5em minus 0.4em\relax {ACM}, 2020, pp. 2307--2316.

\bibitem{AutoDis}
H.~Guo, B.~Chen, R.~Tang, Z.~Li, and X.~He, ``Autodis: Automatic discretization
  for embedding numerical features in {CTR} prediction,'' \emph{CoRR}, vol.
  abs/2012.08986, 2020.

\bibitem{AutoFT}
X.~Yang, Q.~Liu, R.~Su, R.~Tang, Z.~Liu, and X.~He, ``Autoft: Automatic
  fine-tune for parameters transfer learning in click-through rate
  prediction,'' \emph{CoRR}, vol. abs/2106.04873, 2021.

\bibitem{AutoLoss}
X.~Zhao, H.~Liu, W.~Fan, H.~Liu, J.~Tang, and C.~Wang, ``Autoloss: Automated
  loss function search in recommendations,'' \emph{CoRR}, vol. abs/2106.06713,
  2021.

\end{thebibliography}

\end{document}